\documentclass{article}
\usepackage{amsmath}
\usepackage{verbatim}
\usepackage{amssymb}
\usepackage{amsthm}
\usepackage{graphicx}
\usepackage{subfigure}
\usepackage{mathrsfs,mathtools}
\usepackage{bm}
\usepackage{tikz,pgfplots}
\usepackage{parskip}

\usepackage{fullpage}
\newtheorem{theorem}{Theorem}

\theoremstyle{definition}

\usepackage[colorlinks=true,pagebackref,linkcolor=magenta,bookmarks=false]{hyperref}
\usepackage[colon,sort&compress]{natbib}
\newcommand{\argmax}{\operatornamewithlimits{argmax}}
\newcommand{\argmin}{\operatornamewithlimits{argmin}}
\renewcommand{\tilde}{\widetilde}

\newcommand{\Real}{\mathbb{R}}
\newcommand{\GG}{\mathbb{G}}
\newcommand{\EE}{\mathbb{E}}
\newcommand{\mX}{\mathcal{X}}

\newcommand{\mG}{\mathcal{G}}
\newcommand{\mF}{\mathcal{F}}
\newcommand{\iid}{\stackrel{\mathrm{iid}}{\sim}}
\newcommand{\ind}{\stackrel{\mathrm{ind}}{\sim}}
\makeatletter
\def\thm@space@setup{%
   \thm@preskip=\parskip \thm@postskip=0pt
}
 \makeatother

\begin{document}
\title{Statistical inference on errorfully observed graphs}
\author{Carey E. Priebe, Daniel L. Sussman, Minh Tang, and Joshua T. Vogelstein%
\\ Johns Hopkins University, Department of Applied Mathematics and Statistics}

\maketitle

\begin{abstract}

  Statistical inference on graphs is a burgeoning field in the applied
  and theoretical statistics communities, as well as throughout the
  wider world of science, engineering, business, etc.
  In many applications, we are faced with the reality of errorfully
  observed graphs.  That is, the existence of an edge between two
  vertices is based on some imperfect assessment.
  In this paper, we consider a graph $G = (V,E)$.  We wish to perform
  an inference task -- the inference task considered here is ``vertex
  classification'', i.e., given a vertex $v$ with unknown label
  $Y(v)$, we want to infer the label for $v$ based on the graph $G$
  and the given labels for some set of vertices in $G$ not containing
  $v$. However, we do not observe $G$; rather, for each potential edge
  $uv \in {\binom{V}{2}}$ we observe an ``edge feature'' which we use
  to classify $uv$ as edge/not-edge.  Thus we {\it errorfully} observe
  $G$ when we observe the graph $\widetilde{G} = (V,\widetilde{E})$ as
  the edges in $\widetilde{E}$ arise from the classifications of the
  ``edge features'', and are expected to be errorful.
  Moreover, we face a quantity/quality trade-off regarding the
  edge features we observe -- more informative edge features are more
  expensive, and hence the number of potential edges that can be
  assessed decreases with the quality of the edge features.
  We studied this problem by formulating a quantity/quality trade-off
  for a simple class of random graphs model, namely the stochastic
  blockmodel.  We then consider a simple but optimal vertex classifier
  for classifying $v$ and we derive the optimal quantity/quality
  operating point for subsequent graph inference in the face of this
  trade-off. 
  The optimal operating points for the quantity/quality trade-off are surprising and illustrate the issue that methods for intermediate tasks should be chosen to maximize performance for the ultimate inference task.
  Finally, we investigate the quantity/quality tradeoff for errorful observations of the {\it C.\ elegans} connectome graph.

  % The results are surprising ({\color{red} R1 wants more specific}) and suggest that the
  % implications of the quantity/quality trade-off is interesting and
  % non-trivial.

  % We investigate optimal graph inference in the face of this
  % quantity/quality trade-off, and also demonstrate that the optimal
  % choice of edge-classifier for the subsequent graph inference task
  % is {\it not} the Bayes optimal edge-classifier.

\end{abstract}

%Key words: social network; connectome; vertex classification.
\newpage

\section{Introduction}

In areas ranging from connectomics,
where vertices may be neurons and edges indicate axon-synapse-dendrite connections,
%or vertices are brain regions and edges indicate Diffusion Tensor Imaging tracks,
to social networks, where vertices may be people and edges indicate
communication activity, statistical inference on graphs is becoming
essential to scientific, engineering, and business activity. However,
in many of these applications edges cannot be directly observed and
instead we must infer their existence based on auxillary
edge features.  This reality gives rise to errorfully observed graphs,
and the trade-off between more informative but more expensive
edge features and less informative but less expensive edge features is
of fundamental interest.

For example, in connectomics, one often observes image data obtained
from some spatial scanning procedure, and there is a quantity/quality
trade-off between, e.g., spatial resolution of the images and imaging
time (higher resolutions requires longer imaging time) or spatial
resolution and signal-to-noise ratio (higher resolutions implies
lower signal-to-noise ratio per voxel). After the imaging data has
been obtained, a tracing algorithm is then employed on the images to
infer relationships on the brain-graphs. There is once again a
quantity/quality trade-off between how accurate the tracing is and how
many images can be traced. As another example, in social network
analysis, the edges might have attributes associated with them, e.g.,
the text of an email or the voice recording of a telephone call. These
attributes can be quite complex and so procedures such as topic
modeling are commonly used to reduce the edge attribute
complexities. These procedures are often computationally
demanding and thus there is a trade-off in terms of how accurate one
can model the edge attributes and how many edges one can model. See
the Appendix for a more detailed summary expounding upon the relevance
of the quantity/quality trade-off for these two motivating applications.

We investigate optimal graph inference in the face of this quantity/quality trade-off,
and demonstrate that the optimal quantity/quality operating point
can be derived for a surrogate graph inference task.
In the process, we also demonstrate that the optimal choice of edge-classifier
for the subsequent graph inference task
is not necessarily the Bayes optimal edge-classifier.
We also investigate the quantity/quality tradeoff for simulated errorful obesrvations of the {\em C.\ elegans} connectome graph \citep{xu2013computer}.
The {\em C. elegans} is a small worm and the connectome is a representation of the connections between the neurons of the animal as a graph. 
The connectome provides an abstract wiring diagram for how neuron signals can be passed between neurons in the worm.

\subsection{Graph Preliminaries}
\label{sec:graph-preliminaries}

A graph is a pair $G = (V,E)$ with vertices $V = [n] = \{1,\cdots,n\}$
and edges $E \subset {\binom{[n]}{2}}$.  The adjacency matrix $A$ is
$n \times n$, binary, symmetric, and hollow, i.e., the diagonal
entries of $A$ are all $0$; $A_{uv} = 1$ indicates an edge between
vertex $u$ and vertex $v$.

Given a probability space $(\Omega,P)$, a random graph is a graph-valued random variable $\GG: \Omega \to
\mG_n$, where $\mG_n$ denotes the collection of all
$2^{{\binom{n}{2}}}$ possible graphs on $V=[n]$.  A random graph
model, denoted $\mF$, is some specified collection of distributions on
$\mG_n$.  We write $\GG \sim F_{\GG}$ for some distribution $F_{\GG}
\in \mF$.

A simple but interesting random graph model is the stochastic
blockmodel, $\GG \sim SBM([n],B,\pi)$, introduced in
\cite{Holland1983} and of continuing interest
(\cite{Wang1987,Snijders1997Estimation,Airoldi2008}, etc.). Here the
block connectivity probabilities are specified via the $K \times K$
symmetric matrix $B$ with $B_{k_1k_2} \in [0,1]$, and $\pi$ in the
unit simplex $\Delta^K$ specifies the block membership probabilities.
Block membership is given by $Y(v) \iid Multinomial([K],\pi)$, and then
$A_{uv}|Y(u),Y(v) \ind Bernoulli(B_{Y(u),Y(v)})$, yielding independent
edges (conditioned on block membership).

The stochastic blockmodel is motivated by the notion of stochastic
equivalence and community detection. As the probability of an edge
between two nodes depends only on their respective block membership,
two nodes sharing the same block membership are stochastically
equivalent. Nodes with the same block membership can then be
identified as a community, i.e., they share the same communication
patterns. Note that in the stochastic blockmodel, the probabilities of a
connection within blocks is not necessarily larger than those between
blocks. An \emph{affinity} stochastic blockmodel is a stochastic
blockmodel wherein the probabilities of connection within a block is
in general larger than those between blocks. In an \emph{affinity}
stochastic blockmodel, a community is then a collection of nodes whose
connections are more dense within the community, as compared to
connections between that community and other nodes.

A practically useful and theoretically interesting generalization of
the stochastic blockmodel is the latent position model
\citep{hoff02:_laten}.  Consider first fixed latent positions $Z \in
\Real^{n\times d}$, and $\GG \sim LPM(Z,\ell)$ where the link function
$\ell: \Real^d \times \Real^d \to [0,1]$.  Then $A_{uv} \ind
Bernoulli(\ell(Z_{u},Z_{v}))$.  Next, considering random latent
positions, we have $\GG \sim LPM(F,\ell)$, where $Z \sim F$ on
$\Real^{n\times d}$ and $A_{uv}|Z_{u},Z_{v} \ind
Bernoulli(\ell(Z_{u},Z_{v}))$, yielding conditionally (on latent
positions) independent edges. One of the motivation for latent
positions model is the notion of homophily, in which connections
between nodes sharing \emph{similar characteristics} are stronger than
those between nodes sharing different characteristics.  As an example,
in a social network with vertices representing individuals and edges
indicating communications, the latent position of a vertex may be
intepreted as attributes of the individual in the social space, e.g.,
interest in various topics. The communication pattern between
individuals is then determined by their latent positions and the link
function. For example, if the link function is a (monotonic
increasing) radial function such as $\exp(-\|Z_u - Z_v\|^2)$, then
individuals with attributes that are ``close'' together are more likely
to communicate.

A random dot product graph (RDPG) model \citep{young07:_random} is a
special case of the latent position model where the link function is
the inner product and the latent positions are constrained so that
their inner product is always in $[0,1]$; thus
%$\GG \sim RDPG(Z,\langle \cdot,\cdot \rangle)$ or $\GG \sim RDPG(F,\langle \cdot,\cdot \rangle)$.
$RDPG(Z) = LPM(Z,\langle \cdot,\cdot \rangle)$ or $RDPG(F) =
LPM(F,\langle \cdot,\cdot \rangle)$. We note that, as indicated above,
the $K$-block stochastic blockmodel can be view as a special case of a
latent positions model where the number of distinct latent positions
is $K$. For example, take $F$ to be the joint distribution for an
independent sample of size $n$ from a mixture of $d$-dimensional
Dirichlets: $f_{marginal} = \sum_{k=1}^K \pi_k D(r_k \vec{\alpha}_k +
\vec{1})$. Then let block membership be given by $Y(v) \iid
Discrete([K],\pi)$ and latent positions be given by $Z_{v}|Y(v) \ind
D(r_{Y(v)} \vec{\alpha}_{Y(v)} + \vec{1})$.  Finally, let
$A_{uv}|Z_{u},Z_{v} \ind Bernoulli(\langle Z_{u},Z_{v} \rangle)$. This
provides a useful {\it block signal} continuum: when $r_k = 0$ for all
$k$ there is no difference among the blocks, while $\min_k r_k \to
\infty$ yields the $K$-block stochastic blockmodel (when all
$\vec{\alpha}_k$ are distinct).

\subsection{Inference Preliminaries}

\begin{figure}[ht]
\begin{center}
   % \includegraphics[scale=0.375]{YPigraph3.pdf}
   % %\includegraphics[scale=0.325]{cepFIGSsep2012/Laplacian.pdf}
   % \includegraphics[scale=0.375]{YPSTFP.pdf} \\ 
   \includegraphics[width=\textwidth]{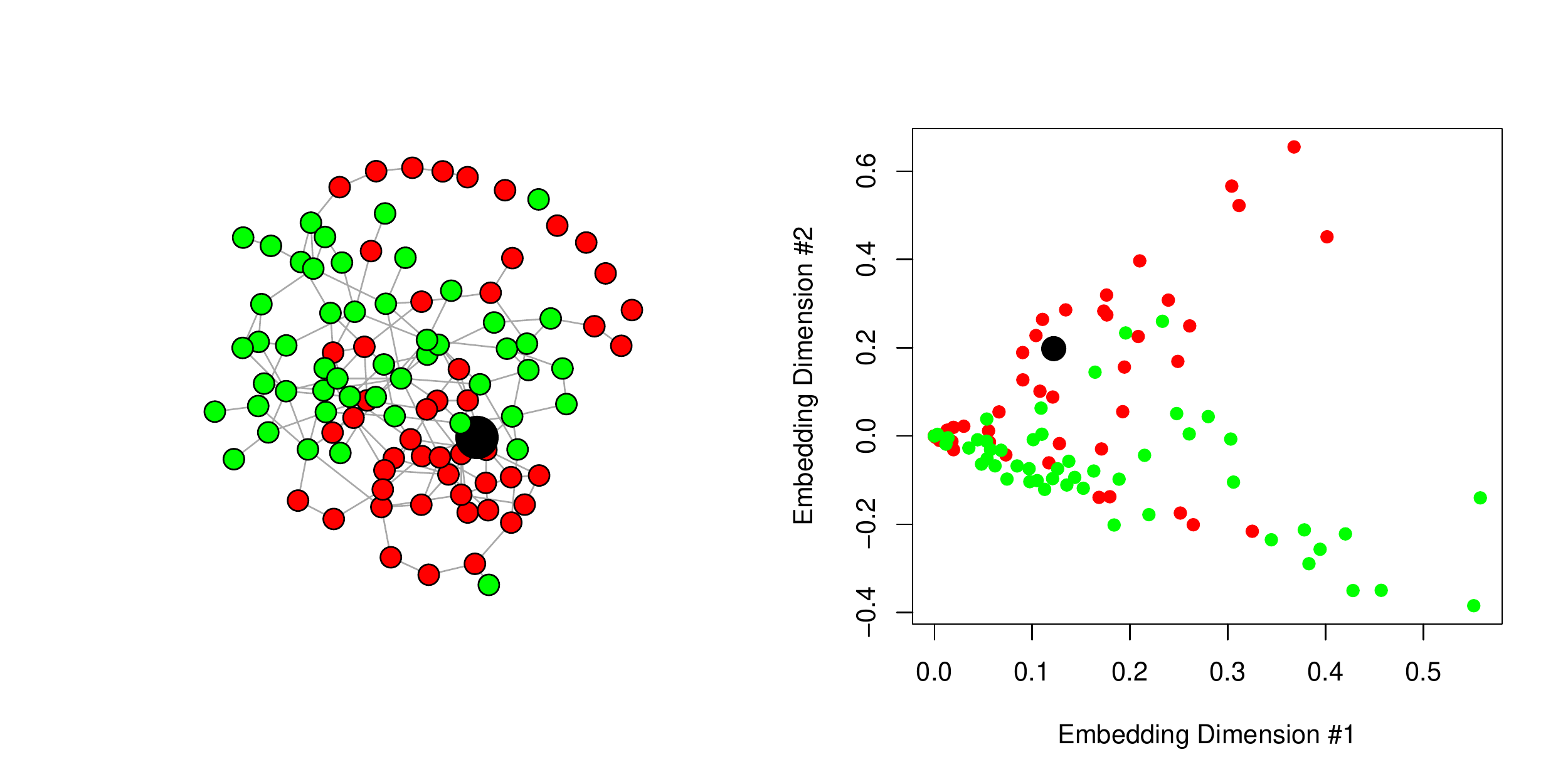}
% Make using figure1() in code/ main.R

   \caption{Illustrative graph inference task: {\it vertex classification}.
   The graph is a simulated random graph distributed according to the stochastic blockmodel with 100 vertices, see \S~\ref{sec:graph-preliminaries}.
Left Panel:
Vertices are labeled as belonging to one of $K=2$ vertex classes -- red and green.
We know these vertex class labels for all but one vertex -- black.
We wish to classify this one unlabeled vertex (based on connectivity structure).
Right Panel:
Once the vertices are embedded in $\Real^2$
(shown here: adjacency-spectral embedding),
the to-be-classified black vertex is easily classified as ``red'' using a $k$-nearest-neighbor classifier.
}
\label{fig:YPi}
\end{center}
\end{figure}

Our goal is graph inference. We may wish to cluster vertices, or
identify important vertices, or merely perform exploratory data
analysis on the graph, looking for interesting structure.  For
concreteness, we assume that vertices are labeled as belonging to one
of $K$ vertex classes (e.g., professors, postdocs, students, etc.)
and that we know these vertex class labels for some subset of
vertices.  In this case, we wish to classify the unlabeled vertices
(based on connectivity structure).  See Figure \ref{fig:YPi}(left).
One common methodology for vertex classification is to embed the graph
into finite-dimensional Euclidean space and employ standard
classification methodologies.  See Figure \ref{fig:YPi}(right).

The embedding depicted in Figure \ref{fig:YPi} is an
adjacency-spectral embedding,  the direct embedding of the adjacency matrix $A$,
which is particularly appropriate for the random dot product graph
model, as considered in \cite{STFP} and \cite{FSTVP}. 
There are many graph embedding
  techniques, with perhaps the most popular being various
  instantiations of the Laplacian eigenmap (see, e.g.,
  \cite{Belkin03laplacianeigenmaps}); we shall not be concerned in
  this paper with the comparative properties of graph embedding
  techniques.
\cite{STP}
demonstrates that $\widehat{Z} = \argmin_{Z \in \Real^{n\times d}} ||A -
ZZ^T||_F$ admits {\it universally consistent classification}
\citep{DGL1996} for random dot product graphs. That is to say, let the rows of $Z
\in \Real^{n\times d}$ be a collection of (latent) positions, with each
row of $Z$ having a class label $k \in \{1,2,\dots,K\}$.  Now let $A$ be the adjacency matrix corresponding to a random dot product
graph generated by $Z$. If we estimate $Z$ by $\widehat{Z}$, then in the limit as $n
\rightarrow \infty$ the classification error based on the estimated
$\widehat{Z}$ can be made as low as the Bayes error rate obtainable
when classifying using the true but unobserved $Z$, for any joint
distribution of the latent positions and the class labels.

However, in this paper there are {\it two} classification tasks to be
considered: vertex classification and edge classification.  The ultimate goal is graph inference.  The
{\it surrogate} inference task considered here is ``vertex
classification''; that is, we consider vertex class labels $Y(v) \iid
Multinomial([K],\pi)$ and attempt to recover the unobserved vertex class
label for distinguished vertex $v^* \in [n]$ based on the observed
vertex class labels for $v \in [n] \setminus \{v^*\}$ and the observed
graph $G=([n],E)$.  In addition, the errorful nature of our graph
observation process induces an edge-classification task; we do not
observe $E$ but rather edge features $X(uv)$ for each potential edge
$uv \in {\binom{[n]}{2}}$ from which we must infer $E$,
and subsequent graph inference depends on this edge-classification.

\subsection{Outline}

In Section 2, we present a model for errorfully observed graphs which
admits investigation of the quantity/quality trade-off. Our model for
errorfully observed graphs is
based on the stochastic blockmodel for random graphs in which edge features are observed rather than edges themselves leading to an edge classification task.
In Section 3,
we develop the framework for vertex classification, a graph inference task. In
Section 4, we demonstrate that the optimal operating point for the
quantity/quality trade-off can be identified for our inference task.
We also demonstrate the same quantity/quantity tradeoff in an investigation of 
errorful observations of the {\em C. Elegans} connectome graph.
We conclude in Section 5 with a discussion of extensions and implications of this
work.

\section{Errorfully Observed Graphs}\label{sec:errGraph}

For each potential edge $uv \in {\binom{[n]}{2}}$ we observe an
$\mX$-valued edge feature $X(uv)$.  These features may be as complex
as ``all the information regarding all interactions between actors $u$
and $v$'' -- for instance, electron microscope imagery of axons and
dendrites for neurons $u$ and $v$ or the text of all emails twixt
adresses $u$ and $v$.  We will assume for simplicity that the $X$'s
take their values in $[0,1]$.  In both connectomics and social
networks, for example, this is often a reasonable assumption:
``Peters' rule'' \citep{6881} suggests that the probability of synapse
is proportional to axon/dendrite proximity; topic models (see
\cite{Blei:2012:PTM:2133806.2133826} for a recent survey) estimate the
proportion of topic ``sports'' (say) for each text document, and then
the graph of interest is ``who talks to whom about sports.''

Each edge feature $X(uv)$ is associated with the true class label for the
potential edge  $Y(uv)$. Here, $Y(uv)=1$ indicates that the edge between
vertices $u$ and $v$ is present while $Y(uv)=0$ indicates its absence. (Note, we
will use $Y$ to denote the class label for {\it both} classification tasks to
be considered; it will be easy to distinguish between $Y(v)$, a class label
associated with a single vertex, and $Y(uv)$, a class label associated with a
pair of vertices, i.e., a potential edge.) The distribution of $X(uv)$ is
governed by the value of $Y(uv)$ so that for $Y(uv)=y\in\{0,1\}$, the class-conditional distributions of the edge features are $F_{X(uv)|Y(uv)=y}=F_y$.
Furthermore, we assume the edge features are iid given  the presence or
absence of an edge so that $X(uv)|Y(uv)=y \iid F_{y}$.  That is, the edge-feature
 distribution for potential edge $uv$ depends on only $Y(uv)$ (edge
/not-edge).
% Additionally, we assume for simplicity that  the edge-feature
%class-conditional distributions have densities denoted $f_0$ and $f_1$.

We will assume that the true potential edge class labels $Y(uv)$ are
unobserved and instead we only observe the edge features $X(uv)$, so rather
than observing the true graph we  observe a collection of edge features for
some potential edges. To facilitate subsequent inference, our goal then is to
estimate the unobserved true graph by classifying potential edges using the
edge features.

For a random graph $\GG \sim F_{\GG}$, let $\rho = \rho(\GG) =
\EE[|E|/{\binom{n}{2}}]$ denote the probability that an arbitrary $uv \in
{\binom{[n]}{2}}$ is an edge in the (random) graph; that is, the expected
graph density.   In this case the edge-feature marginal distribution is $F_{X} =
(1-\rho) F_0 + \rho F_1$. Without regard to the graph setting and using
standard statistical  pattern recognition results,  we can identify the Bayes
edge classifier  based on the edge-feature marginal given by $g_{Bayes}(x) =\argmax_{y\in\{0,1\}} P[Y=y|X=x]$.  This results in the random graph
$\widetilde{\GG}_{Bayes}$, whose distribution is induced by $F_{\GG}$, $F_0$,
and $F_1$.  (NB: Edge classification is not the ultimate goal.  Rather, edge
classification is an enabling step for subsequent (errorful) graph inference.
The optimality of $g_{Bayes}$ for this subsequent inference will be addressed
in Section~\ref{sec:quantQual}.)

We will assume for simplicity that
the $[0,1]$-valued edge features $X_0 \sim F_0$ and $X_1 \sim F_1$ satisfy the
stochastic ordering condition $X_0 <_{ST} X_1$; that is, larger values of the
edge feature $X(uv)$ indicate that the potential edge $uv$ is more likely
truly an edge.  In light of this assumption, we will consider the collection
of edge-classifiers given by $g_{\tau}(X) = I\{X>\tau\}$ for threshold $\tau
\in [0,1]$.  Note that for the simulation considered in \S~\ref{sec:demo}, this collection of classifiers includes the Bayes optimal edge classifier.

However, we also have a quantity/quality trade-off: {\it more   informative}
edge features are {\it more expensive}.  
To capture the idea of our quantity/quality trade-off, we will suppose that there 
are a collection of class-conditional edge-feature distributions some of which are
expensive, so that a meager number of edge features $X(uv)$ will be observed, while 
others
are cheap, so that many or even all of the edge features will be observed.
For expensive edge features, the class-conditional distributions are well separated 
and easy to classify but for cheap edge features the distributions will mostly overlap.
If an edge feature for a potential edge is observed we will say that the potential edge has 
been {\em assessed}.

We index the collection of class-conditional edge-feature distributions
$F_{0,\kappa},F_{1,\kappa}$ with the {\it   quality} index $\kappa \in
(0,\infty)$ such that larger $\kappa$ implies {\it more informative but more
expensive} edge features.  To accomodate the quality/quantity  tradeoff, there
are natural stochastic ordering conditions: (a) $X_{0,\kappa} <_{ST}
X_{1,\kappa}$ for all $\kappa$, so that the classifier $g_\tau$ is reasonable, and (b) $\kappa_1 < \kappa_2$ implies
$X_{0,\kappa_1} >_{ST} X_{0,\kappa_2}$ and $X_{1,\kappa_1} <_{ST}
X_{1,\kappa_2}$, so that higher quality edge features make classifying potential edges more accurate.  Now we introduce the {\it decreasing   quality penalty
function} $h: \Real_+ \to [0,1]$. We actually assess only
$100{\cdot}h(\kappa)\%$ of the potential edges, so that larger $\kappa$
implies {\it more informative but more expensive} edge features and hence
fewer potential edges actually classified. We assume that the potential edges
not  assessed due to the quality penalty $h(\kappa)$ are   Missing
Completely At Random (MCAR).

%$\EE[\rho(\GG)] = ( n \pi^T B \pi - 1^T diag(B) \pi )/(n-1)$ 
We write the
collection of potential edges $uv$ as the disjoint union of edges ($uv \in E
\iff Y(uv)=1$) and non-edges ($uv \in \overline{E} \iff Y(uv)=0$); thus
$\binom{[n]}{2} = E \sqcup \overline{E}$.  If we denote the  set of edges in
the estimated graph by $\widetilde{E}$, i.e. the set of potential edges
assessed and classified as actual edges, then the event $\{ uv \in
\widetilde{E} \}$ depends on $\tau$ through the edge classifier $g_{\tau}$ and
on $\kappa$ and $Y(uv)$ through the class-conditional edge-feature
distribution $F_{Y(uv),\kappa}$.  Given the class-conditional edge-feature
distributions $F_{0,\kappa}$ and $F_{1,\kappa}$ and $\tau \in [0,1]$, 
the probability that a potential
edge that is truly an edge is assessed and correctly classified as an edge is
 \[ P_{\tau,\kappa}\left[ uv \in \widetilde{E} ~ | ~ uv\text{ assessed and}
~ uv \in E \right] = 1-F_{1,\kappa}(\tau).\]
Similarly, the probability that a
potential edge that is truly not an edge is incorrectly classified as an edge is
\[ P_{\tau,\kappa}\left[ uv \in \widetilde{E} ~ | ~ uv\text{ assessed and} ~ uv
\in \overline{E} \right] = 1-F_{0,\kappa}(\tau).\]
We must
also account for the quality penalty $h:(0,\infty) \to [0,1]$, decreasing for
$\kappa \in (0,\infty)$.  Incorporating this penalty, we have
$P_{\tau,\kappa}[uv\in\tilde{E}| uv\in E] = h(\kappa) (1-F_{1,\kappa}(\tau))$
and $P_{\tau,\kappa}[uv\in \tilde{E}|uv\in
\bar{E}]=h(\kappa)(1-F_{0,\kappa}(\tau))$. 
Additionally, we choose to set non-assessed potential edges to be non-edges in the final graph, see below and \S~\ref{sec:missing-model}.

This framework results in the following {\it errorfully observed stochastic
blockmodel}.  Assume that the true underlying graph $\GG \sim SBM([n],B,\pi)$ 
(see \S\ref{sec:graph-preliminaries}).
Using the simple results in the previous paragraph, we define
\begin{equation}   \label{eq:1}  \widetilde{B} =
h(\kappa)\left[(1-F_{1,\kappa}(\tau)) B + (1-F_{0,\kappa}(\tau)) (J-B)\right],
\end{equation} 
where $J$ is the $K \times K$ matrix of all 1's.
The resultant
{\em errorfully observed graph distribution}, i.e. the distribution of the graph
constructed by assessing and classifying a subset of the potential edges, is given by
$\widetilde{\GG} \sim SBM([n],\widetilde{B},\pi)$. Based on Eq.~\eqref{eq:1},
$h(\kappa)$ can be viewed as the probability that a potential edge is assessed
and $1-F_{1,\kappa}(\tau)$ and $1-F_{0,\kappa}(\tau)$ are the true and false
positive rates, respectively, for the edge classification task, which is only
performed on assessed edges.

%We now clarified some details behind our errorful model and the choice
%of edge features.

Another intepretation of the errorful graph model $\tilde{\mathbb{G}} \sim
SBM([n], \tilde{B}, \pi)$ is as follows.  We start with an unobserved
random graph $\mathbb{G}=(V,E)$ distributed according to the stochastic
blockmodel  with parameters $B$ and $\pi$. Based on a selected quality index $\kappa\in(0,\infty)$, for each potential edge $uv$, an
edge feature $X(uv)$, distributed according to $F_{1,\kappa}$ if $uv\in E$ or $F_{0,\kappa}$ if $uv\notin E$, is generated. Let $\mathbb{G}_{f}$ be the collection of all
edge features for each potential, not all of which will be observed.  From
$\mathbb{G}_{f}$, we observe a random subset of the edge features for the assessed potential edges and classify the
potential edges as edges and non-edges via some classification rule $g$ that
depends on the parameter $\tau$ (with the remaining edge features being
automatically classified as non-edges). The number of assessed potential edges can be viewed
as a binomial random variable with $\tbinom{n}{2}$ trials and success
probability $h(\kappa)$. The graph $\tilde{\mathbb{G}}$ resulting from the
edge classifier on the assessed potential edges is the starting point for our vertex classifier
and corresponds to the stochastic blockmodel with parameters $\tilde{B}$ and
$\pi$ as defined in Eq.~\eqref{eq:1}. Note that $\widetilde{\GG}$ and $\widetilde{B}$ will
always depend, implicitly, on $\kappa$ and $\tau$.  This formulation assumes
that the potential edges that are not assessed, due to the quality penalty
$h(\kappa)$, are set to 0 -- i.e., non-edge.  This choice of dealing with
missing values for the potential edges will be revisited later in \S~\ref{sec:missing-model}.

Regarding the edge features, our assumption that they are in the
interval $[0,1]$ is for ease of exposition only. In general, the
features can take values in any arbitrary space, provided that a
reasonable notion/intepretation for stochastic ordering exists in that
space. For example, the features can take values in $\mathbb{R}^{k}$. One can then classify a potential edge $e$ as
an edge if the corresponding feature has value in some subset $S \subset
\mathbb{R}^{d}$. $S$ will then serve the role similar to that of $\tau$ in our current
setup. The notion of stochastic ordering then corresponds to
the notion that the $F_{0,\kappa}(S) \leq F_{1,\kappa}(S)$ for any
$\kappa$, and that for $\kappa < \kappa'$, $F_{0,\kappa'}(S) \leq
F_{0,\kappa}(S) \leq F_{1,\kappa}(S) \leq F_{1,\kappa'}(S)$.

Finally, there is the issue of using an edge classifier to
classify the assessed potential edges before performing vertex classification
versus working directly with the edge features. That is to say,
instead of enforcing dichotomous edge relationships when performing
vertex classification, one can try to explicitly model the edge
features themselves. We will address this in more detail in the next
section, but for now, we remark that the issue of quantity/quality trade-off, as induced by the quality index $\kappa$ and the penalty function
$h(\kappa)$, is independent of our imposition of edge classification as an
intermediary inference task.

\section{Vertex Classification}

%We present a random graph model for errorfully observed
%$\widetilde{\GG}$ and demonstrate, among other things, that
%$g_{Bayes}$ is {\it not}, in fact, the optimal edge classifier
%{\it\bf for subsequent graph inference}.
%Given a graph ${\GG}$ with vertex class labels $Y(v) \iid Discrete([K],\pi)$,

Given a graph $G=([n],E)$ with vertex class labels $Y(v) \in [K]$, there are
many methodologies available for estimating the unobserved vertex class label
for a distinguished vertex $v^* \in [n]$ based on the observed vertex class
labels for $v \in [n] \setminus \{v^*\}$ (recall Figure \ref{fig:YPi}).  We
will proceed with perhaps the simplest nontrivial vertex classification
approach; later, we will see that we can optimize this classifier for $\tau$
and $\kappa$ in the errorfully observed stochastic blockmodel. Briefly, each
vertex $v$ is classified as belonging to the block such that $v$ is most
likely to be adjacent to vertices in that block.

First, for each $k \in [K]$, we count the number of class $k$ vertices
$n_k = \sum_{v \in [n] \setminus \{v^*\}} I\{Y(v)=k\}$.  Next, we
calculate the $k$-degree of $v^*$ -- the number of class $k$ vertices
that are connected to $v^*$ -- given by $d_k(v^*) = \sum_{v \in [n]
\setminus \{v^*\}} I\{Y(v)=k\} \cdot I\{vv^* \in {E}\}$.  Finally, we
classify $v^*$ via $\gamma(v^*) = \argmax_k d_k(v^*)/n_k$.

The classifier $\gamma$ makes perfect sense for an {\it affinity}
stochastic blockmodel -- that is, a stochastic blockmodel with
$B_{kk} > B_{kk'}$ for each $k$ and for all $k' \neq k$: assume that
$\GG \sim SBM([n],B,\pi)$ with $B$ satisfying the affinity conditions
and that $v^*$ is chosen uniformly at random, and see that
$D_k(v^*)/N_k \approx B_{kY(v^*)}$. Here we have
written $D_k(v^*)$ and $N_k$ for the random variable versions of
$d_k(v^*)$ and $n_k$ defined above.

In fact, $\gamma$ is even the
Bayes-optimal classifier in our $2 \times 2$ setting for $B$ with
$B_{11} = B_{22}$, $B_{12} = B_{21} < B_{11}$ and $n_1=n_2$.
Indeed, under this setting, the simple vertex classifier corresponds
to the likelihood ratio test. This can be seen as follows. For this model, since the block membership probabilities are equal, the Bayes classifier for classifying a vertex $v$ as class $1$ or
$2$ depends on whether the likelihood ratio
\begin{equation*}
  \frac{ \tbinom{n_1}{D_1(v^*)} B_{11}^{D_1(v^*)} (1 - B_{11})^{n_1 - D_1(v^*)} \tbinom{n_2}{D_2(v^*)}
    B_{12}^{D_2(v^*)} (1 - B_{12})^{n_2 - D_2(v^*)}}  {\tbinom{n_1}{D_1(v^*)} B_{21}^{D_1(v^*)} (1 - B_{21})^{n_1 - D_1(v^*)} \tbinom{n_2}{D_2(v^*)}
    B_{22}^{D_2(v^*)} (1 - B_{22})^{n_2 - D_2(v^*)}}
\end{equation*}
is greater than or less than $1$, where $D_1(v^*)$ is the number of
vertices in class $1$ adjacent to $v^*$ and $D_2(v^*)$ is the number of
vertices in class $2$ adjacency to $v^*$.
Some algebraic simplifications
lead to checking whether
\begin{equation*}
  \Bigl(\frac{B_{11}}{B_{12}}\Bigr)^{D_1(v^*) - D_2(v^*)}
  \Bigl(\frac{1 - B_{12}}{1 - B_{11}}\Bigr)^{D_1(v^*) - D_2(v^*)}
\end{equation*}
is greater than or less than $1$. This is equivalent to checking
whether $D_1(v^*) \geq D_2(v^*)$ as $B_{11} > B_{12}$ implies $1 - B_{11} \leq
1 - B_{12}$.

Define
$L = P[\gamma(v^*) \neq Y(v^*)|{\GG},\{Y(v)\}_{v \in [n] \setminus \{v^*\}}]$
to be the probability of misclassifying vertex $v^*$ using classifier $\gamma$
\citep{DGL1996}.
A simple conditioning argument yields the following result.

\begin{theorem}
\label{thm1}
Let $\GG \sim SBM([n],B,\pi)$ be an affinity stochastic blockmodel
graph.  Let the classifier $\gamma(v^*) = \argmax_k D_k(v^*)/N_k$.
Conditional on $[N_1,\cdots,N_K] = [n_1,\cdots,n_K]$ and $Y(v^*) = k$,
the binomials $D_{1}(v^*)\sim Bin(n_1,B_{k1})$, $\cdots$, $D_{K}(v^*)\sim Bin(n_K,B_{kK})$ are
independent.  Thus the probability of misclassification {\it with no
  ties} in the classification rule is given by $P[D_k/n_k
< \max_{k' \neq k} Bin(n_{k'},B_{kk'})/n_{k'}]$; the probability of
misclassification {\it in the case of ties}, given by $T_k$, depends
on the tie-breaking procedure.  Therefore, the probability of
misclassification is given by
\begin{eqnarray}
  \label{eq:2}
L &=&  P[\gamma(v^*) \neq Y(v^*)|{\GG},\{Y(v)\}_{v \in [n] \setminus \{v^*\}}] \nonumber \\
   %&=&  \sum_{n_1 + \cdots + n_K = n-1} { {n-1} \choose {n_1, \cdots, n_K} } \pi_1^{n_1} \cdots \pi_K^{n_K}
&=&  \sum_{\mathscr{S}} { \binom{n-1}{n_1, \cdots, n_K} } \prod_{k=1}^K \pi_k^{n_k}
\sum_{k=1}^K \pi_k \left( P\left[\frac{Bin(n_k,B_{kk})}{n_k} < \max_{k' \neq k} \frac{Bin(n_{k'},B_{kk'})}{n_{k'}}\right] + T_k \right)
%&=&  \sum_{\mathscr{S}} { {n-1} \choose {n_1, \cdots, n_K} } \prod_{k=1}^K \pi_k^{n_k}
%\sum_{k=1}^K \pi_k \left( P\left[\frac{D_{k}(v^*)}{n_k} < \max_{k' \neq k} \frac{D_{k'}(v^*)}{n_{k'}} | Y(v^*) =k, N_1=n1,\dotsc,N_K=n_k\right] + T_k \right)
\end{eqnarray}
where the first summation with subscript $\mathscr{S}$, for the multinomial, is over
all non-negative integer partitions of $n-1$ into $[n_1,\cdots,n_K]$.
(The convention $\frac{0}{0}=0$ must be adopted for the cases in which some $n_k$ are 0.)
\end{theorem}
Note that in Theorem~\ref{thm1} we assume that the stochastic blockmodel graph $\mathbb{G}$ is perfectly observed and so $\tau$ and $\kappa$ do not enter in to the calculations.
In the next section we optimize the classifier $\gamma$ for $\tau$ and $\kappa$
in the errorfully observed affinity stochastic blockmodel.

% Dan's Remark
%\begin{remark}
We now remark on our setup where we classify the
potential edges into edges and non-edges based on the edge features. It is possible, and even
perfectly reasonable, to explicitly model the edge features instead of
classifying or assuming a dichotomous edge relationship as we do in
this paper. Furthermore, assuming that our model for the edge features
is accurate, it is also possible to identify the Bayes optimal vertex
classifier. Suppose that for the to-be-classified vertex $v^*$, we
observe $m_k$ edge features for potential edges from $v^*$ to vertices
in block $k$.  Denote, these edge features by $X= \{ X^{(k)}_{i}:
k\in[K],i \in [m_k]\}$.  Given $B$, $\pi$ and densities for edge
features $f_{1,\kappa}$ and non-edge features $f_{0,\kappa}$, the optimal vertex
classifier is given by
\begin{align*}
    g^*(X) &= \argmax_{y\in[K]} P[Y(v^*)=y|X] \\
    &= \argmax_{y\in[K]} \pi_y \prod_{k=1}^K \prod_{i=1}^{m_k} \left(  B_{yk}f_{1,\kappa}(X^{(k)}_i)+(1-B_{yk}) f_{0,\kappa}(X^{(k)}_i) \right).
\end{align*}
Note that even using this Bayes optimal vertex classifier we
experience a quality/quantity trade-off.  Indeed, as we increase
$\kappa$, the quality of the edge features increases so that the
difference between $f_{0,\kappa}$ and $f_{1,\kappa}$ increases.  On the other hand the
number of observed edge features will decrease, so that $m_k$ will
decrease.  Finding the optimal quality/quantity trade-off in this case,
which is parametrized by only $\kappa$, is also of interest.

However, in many realistic situations it will not be possible to
observe the edge features directly and we will only have access to
pre-classified edge features.  In this situation, in addition to
$\kappa$, there will also be a parameter $\tau$ indexing the
classification of the edge features. In addition, $\kappa$ and $\tau$
must be chosen in advance.  In the example of tracing connections
between neurons in the brain this corresponds to pre-selecting a
threshold for the tracer stopping criterion.  If no such threshold is
selected in advance, it is not clear how edge features would be
represented.  Hence, even though the classifier $g^*$ has minimal error rate 
and so by the information processing inequality will perform at least as 
well as the simple classifier $\gamma$, we choose to investigate the
simpler classifier in order to understand inference in these realistic
scenarios. That is to say, the dichotomous nature of our edge
classification is but a simplifying notion but we believe that whether 
we make this simplification or not, the quantity/quality trade-off phenomena 
to be demonstrated in the next section will be present. We can
therefore assume that they are dichotomous without affecting the core
message of this paper.

\section{Optimizing the Quantity/Quality Trade-Off}\label{sec:quantQual}

Let $\widetilde{\GG} \sim SBM([n],\widetilde{B},\pi)$ be the errorfully observed graph after edge assessment and classification as defined in Section~\ref{sec:errGraph}.  Recall that
the distribution of $\widetilde{\GG}$ depends on the original block
connectivity probability matrix $B$ and on $\kappa$ and $\tau$ (and
hence on the quality penalty function $h$ and on the class-conditional
edge-feature distributions $F_{0,\kappa}$ and $F_{1,\kappa}$),
although this has been suppressed notationally.  Notice also that if
$SBM([n],{B},\pi)$ is an affinity stochastic blockmodel graph, then
so is $SBM([n],\widetilde{B},\pi)$, because the edge-feature
distribution for potential edge $uv$ depends on only $Y(uv)$
(edge/not-edge) and {\it does not otherwise depend on the block
memberships $Y(u)$ and $Y(v)$}.

%NOOOOO
%given an errorfully observed graph $\widetilde{\GG} \sim SBM([n],\widetilde{B},\pi)$ --
%edge-classification has already been performed --
%we move now to the subsequent inference task of vertex classification.

Define $L_{\kappa,\tau} = P[\gamma(v^*) \neq
Y(v^*)|\widetilde{\GG},\{Y(v)\}_{v \in [n] \setminus \{v^*\}}]$ to be
the probability of misclassifying vertex $v^*$ using classifier
$\gamma$ for $\widetilde{\GG}$ with a fixed $\kappa$ and $\tau$.  Eq.\
(\ref{eq:2}) applies, replacing the binomial parameters $B_{k_1k_2}$
with $\widetilde{B}_{k_1k_2}$.  Thus the optimal (quality penalty
parameter $\kappa$, edge-classification threshold $\tau$) pair for the
subsequent vertex classification graph inference problem using
classifier $\gamma$ is given by

\begin{equation}\label{cor1}
(\kappa^*,\tau^*) =
\argmin_{\kappa,\tau} \sum_{\mathscr{S}} { \binom{n-1}{n_1, \cdots,
    n_K} } \prod_{k=1}^K \pi_k^{n_k} \sum_{k=1}^K
\pi_k \left( P\left[\frac{Bin(n_k,\widetilde{B}_{kk})}{n_k} < \max_{k'
\neq k} \frac{Bin(n_{k'},\widetilde{B}_{kk'})}{n_{k'}}\right] + T_k
\right),
\end{equation}
where we recall  the implicit dependence of $\widetilde{B}$ on $\kappa$ and $\tau$ (see Eq.~\eqref{eq:1}).
For the purpose of this paper we do not propose methodology to find the optimal pair $(\kappa^*,\tau^*)$ in Eq~\eqref{cor1}.
In the next section we find an approximate optimal solution via grid search.

\subsection{Demonstration}\label{sec:demo} 

Here we present a simple but illustrative demonstration of  finding the minimizer in Eq.\
\eqref{cor1} and the inherent trade-offs of our model.  Let $SBM([n],{B},\pi)$ be a stochastic blockmodel with
$K=2$, $\pi = [1/2,1/2]^{\prime}$ and $B$ satisfying $1>B_{11}=B_{22}
> B_{12}=B_{21}>0$.  Note that $B$ satisfies the affinity SBM
conditions.  We let the class-conditional edge-features be governed by
Beta distributions: $F_{0,\kappa} = \beta_{2,\kappa}$ and
$F_{1,\kappa} = \beta_{\kappa,2}$. Our choice of a $F_{0,\kappa}$ 
and $F_{1,\kappa}$
is primarily to provide a simple illustrative example and because for $\kappa \in
(2,\infty)$ these distributions satisfy our stochastic ordering
conditions, and that $\kappa=2$ yields useless features and larger
$\kappa$ yields more informative features.  
The particular choice of edge features distributions is relevant only in their subsequent impact on the probabilities of error and further down the line the resulting matrix of edge probabilities $\widetilde{B}$.
Our choice of Beta distributions provides one possible sweep of these parameters that illustrate the quantity/quality tradeoff phenomenon.

Recall that the collection
of edge-classifiers considered is given by $g_{\tau}(X) = I\{X>\tau\}$
for $\tau \in [0,1]$, and notice that $\pi = [1/2,1/2]$ and $B$ doubly
stochastic implies that the expected graph density $\rho(\GG) = ( n
\pi^T B \pi - 1^T diag(B) \pi )/(n-1) = ((n/2)-O(1))/(n-1) \approx 1/2$
and hence, since $f_{0,\kappa}$ and $f_{1,\kappa}$ are reflections
about 1/2 of one another, $\tau_{Bayes} \approx 1/2$ for all $\kappa$.
The quality penalty function considered is $h(\kappa) = (2/\kappa)^3$,
so $\kappa=2$ yields classification of all edges, and while larger
$\kappa$ yields more informative edge features, fewer edges are
actually classified.  We consider $\widetilde{\GG} \sim
SBM([n],\widetilde{B},\pi)$ to be the associated errorfully observed
graph (again, depending on $\kappa$ and $\tau$).  For further
simplicity we condition on $N_1=N_2=(n-1)/2$.

For this demonstration the classifier $\gamma$ simplifies,
yielding \begin{equation}
  \gamma(v^*) = \argmax_k D_k(v^*) = 1 + I\{ D_2(v^*) >
D_1(v^*) \}\label{eq:gammaBayes}
\end{equation} with $D_k(v^*) \ind Bin(n_k,\widetilde{B}_{Y(v^*),k})$.
We can simplify the probability of misclassifcation  error  rate $L_{\kappa,\tau}$ by noting that by conditioning on $N_1=N_2$, the first sum in Eq.~\eqref{eq:1} degenerates into a single summand leaving only the inner sum.
Furthermore, using $\pi_1=\pi_2$ and $\tilde{B}_{1,1}=\tilde{B}_{2,2}$, the probability of misclassification $L_{\kappa,\tau}$ simplifies
to
\begin{align}
 L_{\kappa,\tau} &= \sum_{k=1}^K
\pi_k \left( P\left[\frac{Bin(n_k,\widetilde{B}_{kk})}{n_k} < \max_{k'
\neq k} \frac{Bin(n_{k'},\widetilde{B}_{kk'})}{n_{k'}}\right] + T_k
\right)\notag\\
&= P\left[\frac{Bin(n_1,\widetilde{B}_{11})}{n_1} < \frac{Bin(n_{2},\widetilde{B}_{12})}{n_{2}}\right] + (1/2) P\left[\frac{Bin(n_1,\widetilde{B}_{11})}{n_1} = \frac{Bin(n_{2},\widetilde{B}_{12})}{n_{2}}\right] \notag\\
&= \sum_{i=1}^{n_1} P\left[Bin(n_1,\widetilde{B}_{11}) < i, Bin(n_{2},\widetilde{B}_{12}) = i\right] + (1/2) P\left[Bin(n_1,\widetilde{B}_{11}) = i,  Bin(n_{2},\widetilde{B}_{12})=i\right] \notag\\
&=\sum_{i=1}^{n_1} f_{Bin}(i;n_1,\widetilde{B}_{1,2}) F_{Bin}(i-1;n_1,\widetilde{B}_{1,1}) +
(1/2)\sum_{i=0}^{n_1} f_{Bin}(i;n_1,\widetilde{B}_{1,2}) f_{Bin}(i;n_1,\widetilde{B}_{1,1}). \label{eq:simpleL}
\end{align}
Here, with $K=2$ and conditioning on $N_1=N_2$,
the sensible tie-breaking procedure ``flip a fair coin''
is explicitly accounted for in the second sum in our expression for $L_{\kappa,\tau}$.

\begin{figure}[htbp]
\begin{center}
\includegraphics[scale=0.5]{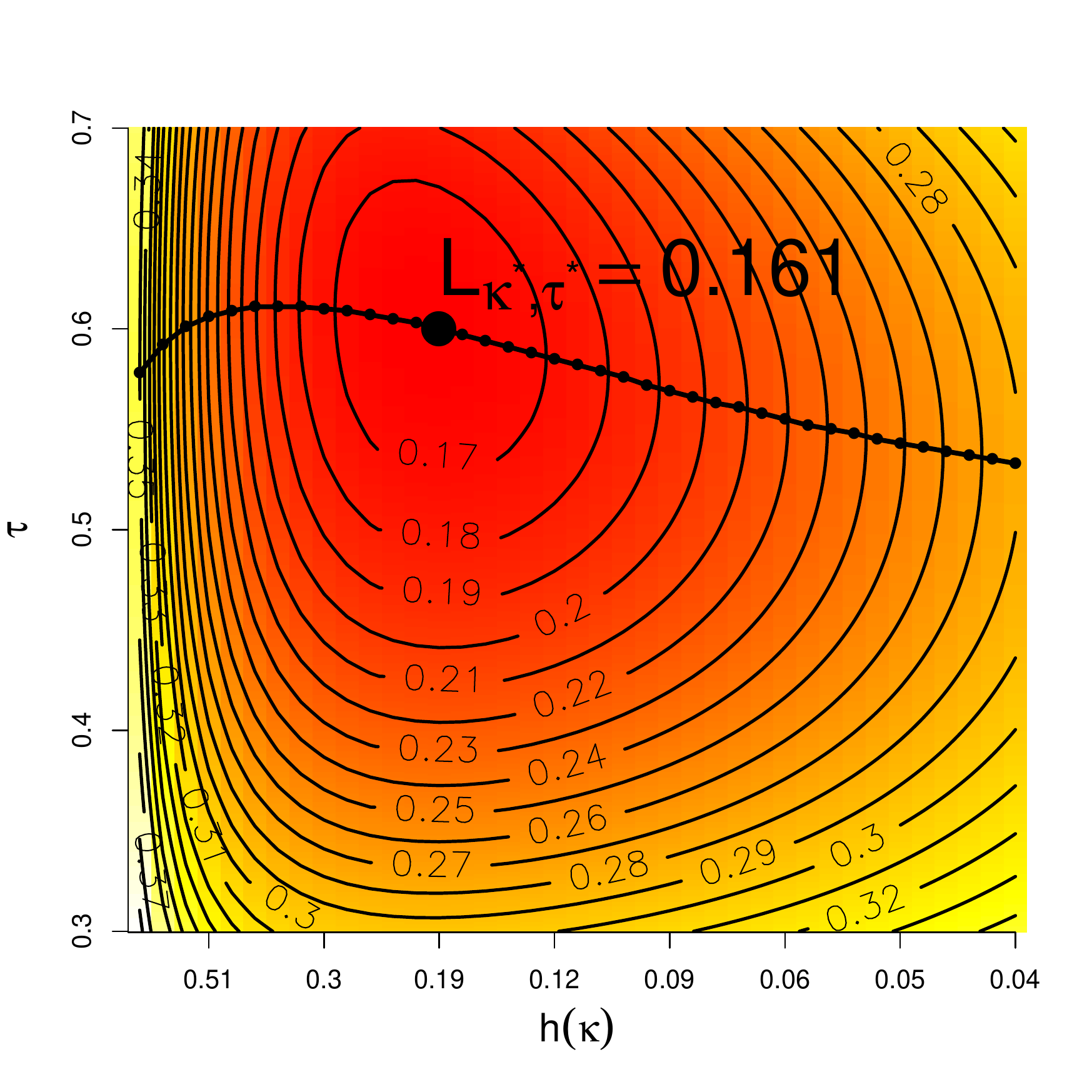}
\caption{Demonstration of optimal inference for errorfully observed graphs:
 $(\kappa^*,\tau^*) \approx (3.5,0.6)$ and $L_{\kappa^*,\tau^*} \approx 0.161$.
 See Section \ref{sec:demo} for details.
 The color and level curves indicate the error rate for the vertex classification task as specified in Eq.~\eqref{eq:simpleL} with the relationship between $\widetilde{B}$ there and $\tau$ and $h(\kappa)$ being specified in Eq.~\eqref{eq:1}.
 The model parameters for the original SBM are $n=51$, $n_1=25=n_2$, $B_{11}=B_{22}=0.9$ and
$B_{12}=B_{21}=0.1$.
 The dotted black curve indicates the optimal classification threshold $\tau$ for each $h(\kappa)$, ie for each $\kappa$.}
\label{fig:Lopt}
\end{center}
\end{figure}

Figure \ref{fig:Lopt} depicts the error surface $L_{\kappa,\tau}$ for this
demonstration, with $n=51$, $B_{11}=B_{22}=0.9$ and $B_{12}=B_{21}=0.1$.  The
$z$-axis -- probability of misclassification $L \in [0,1]$, depicted via color
and level curves -- represents performance on our vertex classification task
computed using Eq.~\eqref{eq:simpleL}.   The $y$-axis -- $\tau \in [0,1]$ --
represents the threshold for the edge classifier used to obtain
$\widetilde{E}$.  The $x$-axis -- $h(\kappa) \in [0,1]$ -- represents the
proportion of assessed edges as a function of the quality $\kappa\in[2,\infty]$ of the edge-features we observe -- larger $\kappa$ implies {\it more
informative but more expensive} edge-features and hence fewer potential edges
actually classified. For this case, $(\kappa^*,\tau^*) \approx (3.5,0.6)$ and
$L_{\kappa^*,\tau^*} \approx 0.161$.

The figure represents this quantity/quality trade-off, and also
demonstrates that the optimal choice of edge classifier is {\it not}
the Bayes optimal classifier ($\tau_{Bayes} \approx 1/2$ for all
$\kappa$).  Indeed, using $\tau_{Bayes}$ rather than $\tau^*$ results
in a substantial relative performance degradation of more than 10\%,
from $L_{\kappa^*,\tau^*} \approx 0.16$ to $\min_{\kappa}
L_{\kappa,\tau_{Bayes}} \approx 0.18$. 

\begin{figure}[htbp]
\begin{center}
   \includegraphics[width=.45\textwidth]{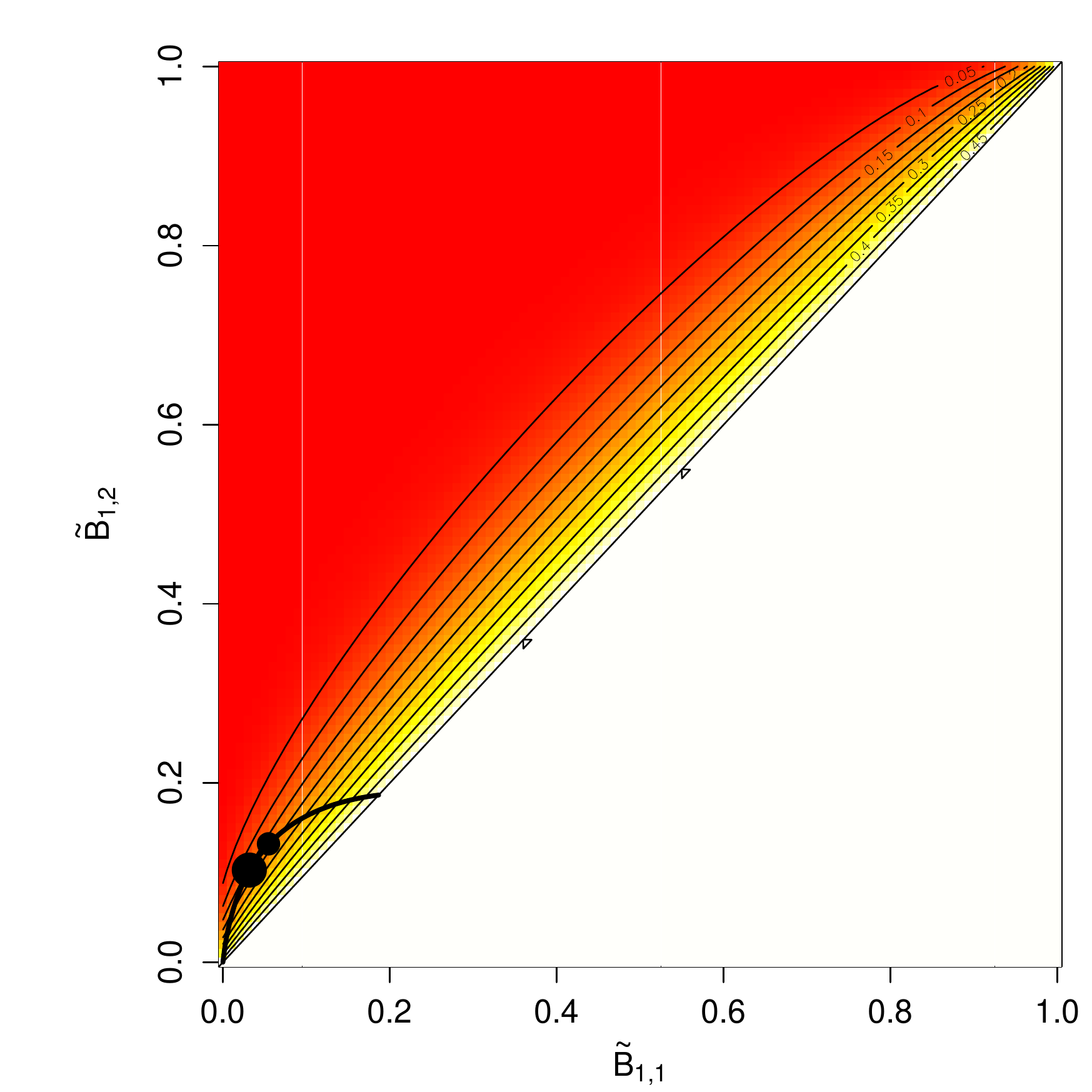}
   \includegraphics[width=.45\textwidth]{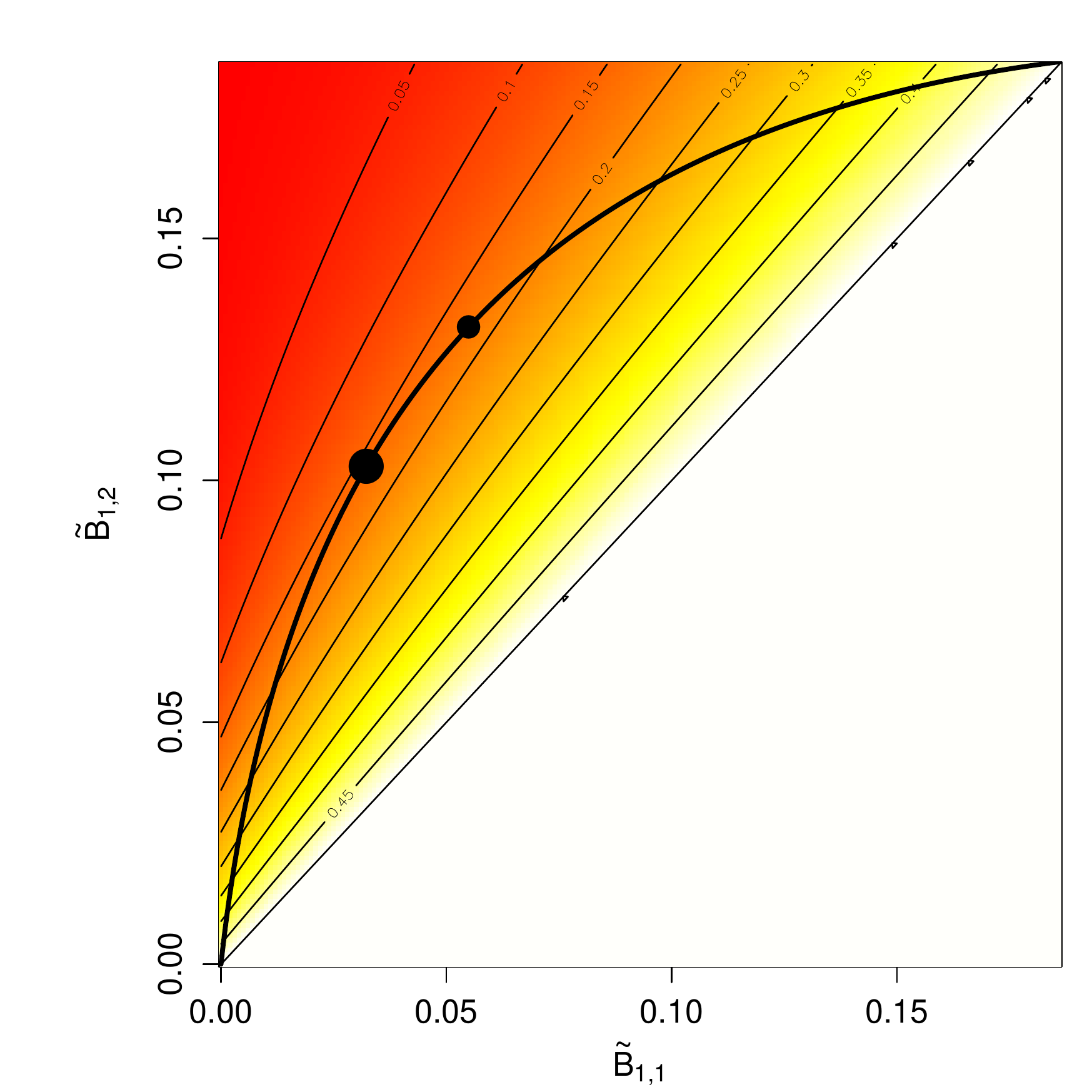}
   \caption{This figure transforms Figure~\ref{fig:Lopt} to show $L_{\kappa,\tau}$ as a function of $\widetilde{B}_{11}$ and $\widetilde{B}_{12}$.
   Again, the color and level curves indicate the error rate for the vertex classification task as defined in Eq.~\eqref{eq:simpleL} as a function of $\widetilde{B}_{11}$ and $\widetilde{B}_{12}$.
   This figure shows the error rate in the $\widetilde{B}_{11},\widetilde{B}_{12}$ coordintes directly.
    The black path shows how $(\widetilde{B}_{11},\widetilde{B}_{12})$, whose explicit dependence on $\kappa,\tau$ is given by Eq.~\eqref{eq:1}, vary as $\tau$ varies from 0 to 1 for
fixed $\kappa=\kappa^*=3.5$.  The axes
represent possible edge presence probabilities for the two blocks in the errorfully observed graph which are also the possible parameter values for the two binomials in the
simplified expression for $L_{\kappa,\tau}$ (see Eq.~\eqref{eq:simpleL}).  The color and level curves represent $L_{\kappa,\tau}$.
The left panel is the full parameter space; the right panel is a
zoom-in of the $(\widetilde{B}_{12},\widetilde{B}_{11})$-path.  This
figure illustrates why the optimal $\tau^*$ (the big black dot) $\neq$
$\tau_{Bayes}$ (the little black dot): the curvature of the
$(\widetilde{B}_{11},\widetilde{B}_{12})$-path does not match the
curvature of the level curves of $L_{\kappa,\tau}$.  }
\label{fig:path}
\end{center}
\end{figure}

\begin{figure}[htbp]
  \centering
  \includegraphics[width=0.5\textwidth]{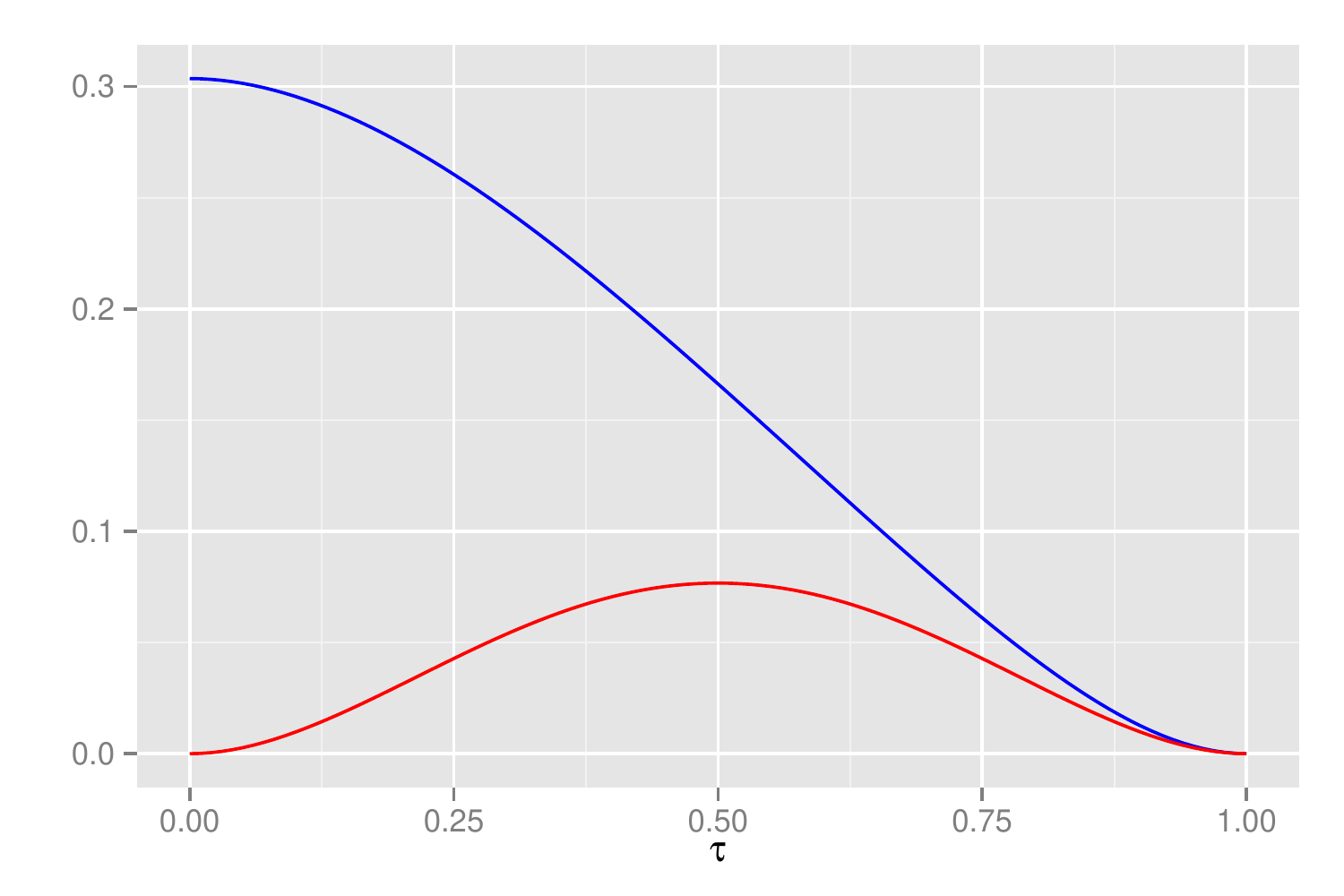}
  \caption{Mean (red curve) and variance (blue curve) of $\tfrac{1}{n}(Z_{1,\tau} -
Z_{2,\tau})$ where $Z_{1,\tau} \sim \mathrm{Bin}(n,
\widetilde{B}_{11})$ and $Z_{2,\tau} \sim \mathrm{Bin}(n,
\widetilde{B}_{12})$ as $\tau$ varies in $[0,1]$. $Z_{1,\tau} -
Z_{2,\tau}$ corresponds to $D_{1}(v^{*}) - D_{2}(v^{*})$ for the
simple vertex classifier $\gamma$.
This figure provides another illustration for why $\tau^*\neq \tau_{Bayes}=1/2$.}
  \label{fig:mean-variance}
\end{figure}

Figure \ref{fig:path} and
\ref{fig:mean-variance} explain this phenomenon by examining the
$(\widetilde{B}_{1,2},\widetilde{B}_{1,1})$-path for fixed
$\kappa=\kappa^*$ as $\tau$ varies from 0 to 1. In
Figure~\ref{fig:path}, we plot the values of $(\widetilde{B}_{12},
\widetilde{B}_{11})$ as $\tau$ varies.
Again, the $z$-axis, depicted with color and level curves, represents the error rate for the vertex classification task as a function $\widetilde{B}$, as in Eq.~\eqref{eq:simpleL}, rather than pulling back to the $(h(\kappa),\tau)$ coordinates as was done in Figure~\ref{fig:Lopt}.
This figure shows how the particular stochastic blockmodel parameters $\widetilde{B}$
for the observed graph impacts performance for the vertex classification task and in comparison to how $\tilde{B}$ varies as a function of $\tau$.

In
Figure~\ref{fig:mean-variance}, we plot the mean and variance of
$\tfrac{1}{n}(Z_{1,\tau} - Z_{2,\tau})$ where $Z_{1,\tau} \sim
\mathrm{Bin}(n, \widetilde{B}_{11})$ and $Z_{2,\tau} \sim
\mathrm{Bin}(n, \widetilde{B}_{12})$ as $\tau$
varies. Figure~\ref{fig:mean-variance} indicates that while the mean
of $Z_{1,\tau} - Z_{2,\tau}$ is unimodal on $[0,1]$ with its largest
value at $\tau = \tfrac{1}{2}$, the variance of $Z_{1,\tau} -
Z_{2,\tau}$ is monotonically decreasing in $\tau$. If
we consider the terms $1-F_{1,\kappa}$ and $1-F_{0,\kappa}$ in Eq.~\eqref{eq:1}
as being mixture coefficients for the matrix of true edges and the
matrix of false edges, then
$\tfrac{1-F_{1,\kappa}(\tau)}{1-F_{0,\kappa}(\tau)}$ increases as $\tau$
increases. That is, even though $\tau = 1/2$ might be the
Bayes-optimal edge classifier for discerning edges, it turns out that
for larger $\tau$, the ratio of true edges among potential edges
(vertex pairs) that are labeled edges also increases. Put another way,
for $\tau = 1/2$, the absolute difference between the number of true edges and
the number of false edges is maximized while for $\tau > 1/2$, though the number
of total edges is smaller, the ratio of true edges to false edges
is larger.

\subsection{Example: {\em C.\ elegans} Connectome}

\begin{figure}[ht]
  \begin{center}
    \includegraphics[width=.75\textwidth]{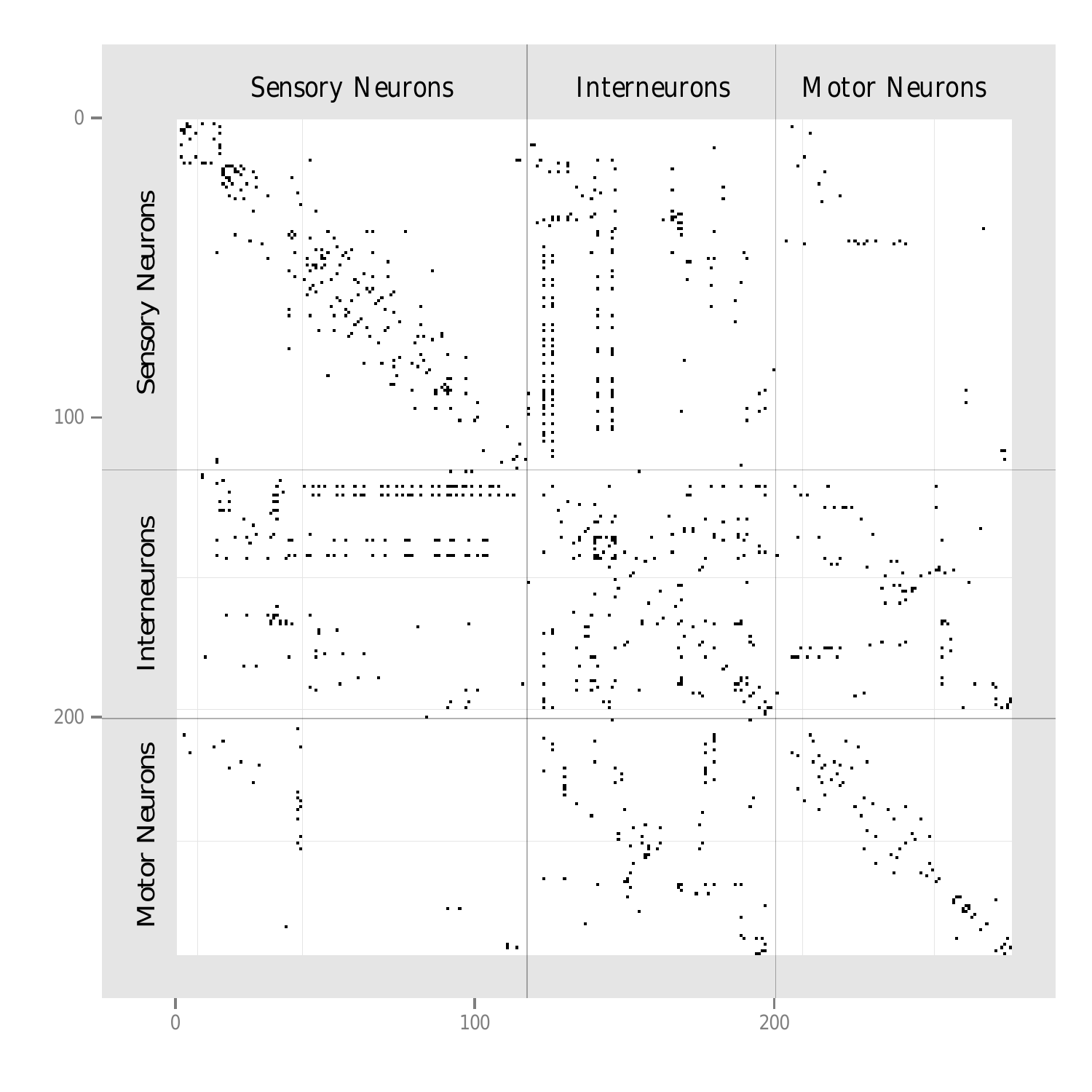}
  \end{center}
  \caption{The adjacency matrix for the {\it C.\ elegans} connectome. The graph has 279 vertices and 515 edges.  Black pixels indicate the presence of an edge and white pixels indicate no edge is present. The vertical and horizontal lines divide the adjacency matrix according to the three types of neurons.}
  \label{fig:CEladj}
\end{figure}

\begin{figure}[ht]
  \begin{center}
    \includegraphics[width=\textwidth]{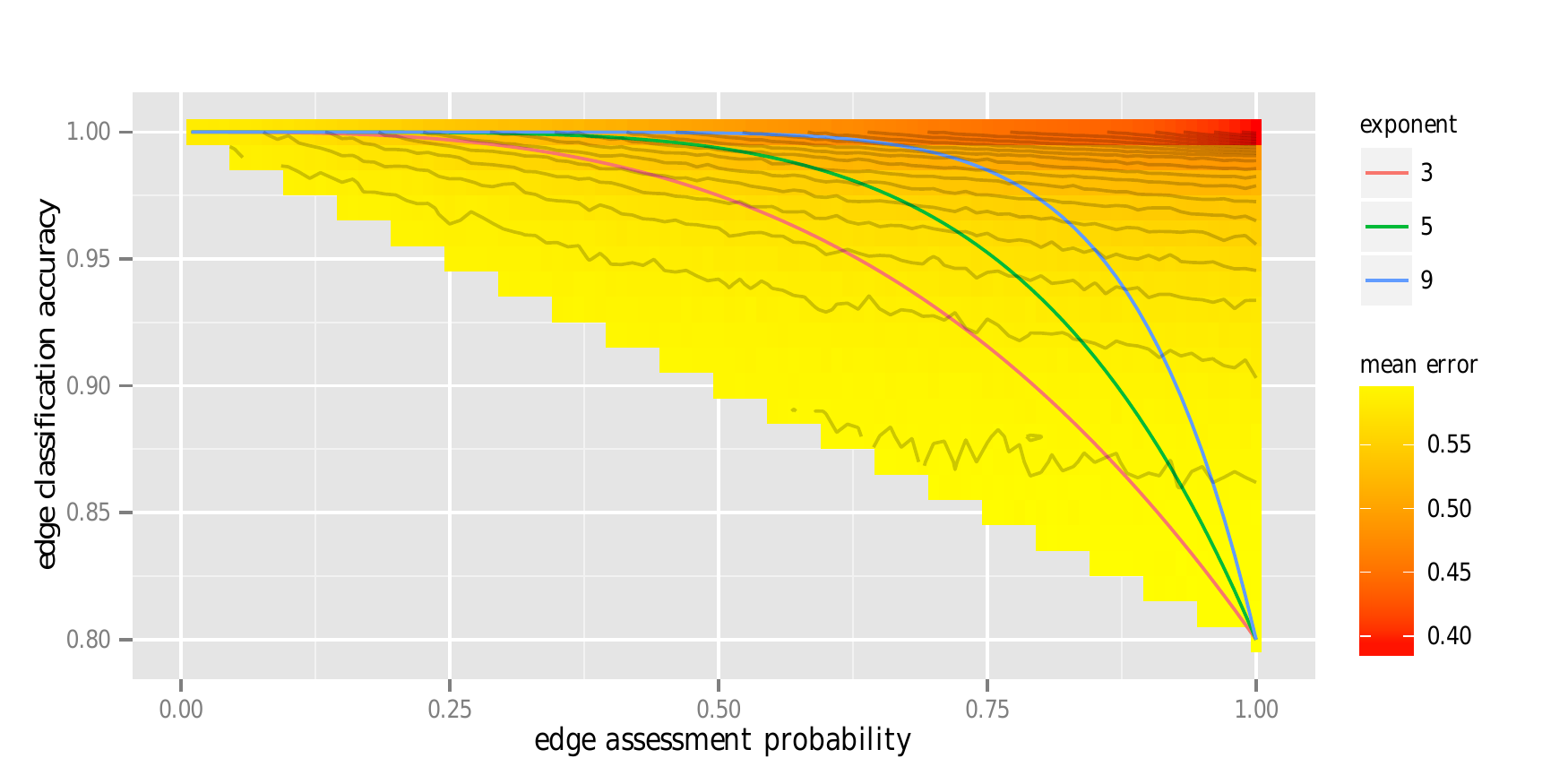}
  \end{center}
  \caption{Each ``pixel'' in the figure corresponds to one selection of the edge assessment probability and edge accuracy probability for an errorful observation of the {\em C. elegans} connectome.
  The  color indicates the mean leave-one-out classification error rate based on 1000 Monte Carlo replicates using the SBM Bayes plugin classifier.
As expected, the error rate decreases substantially as both parameters increase.
The three curves depict three possible profiles for a quality/quantity trade-off between the two parameters.
Figure~\ref{fig:CElcurve} shows the mean error rates along these three curves.}
  \label{fig:CEltri}
\end{figure}

\begin{figure}[ht]
  \begin{center}
    \includegraphics[width=\textwidth]{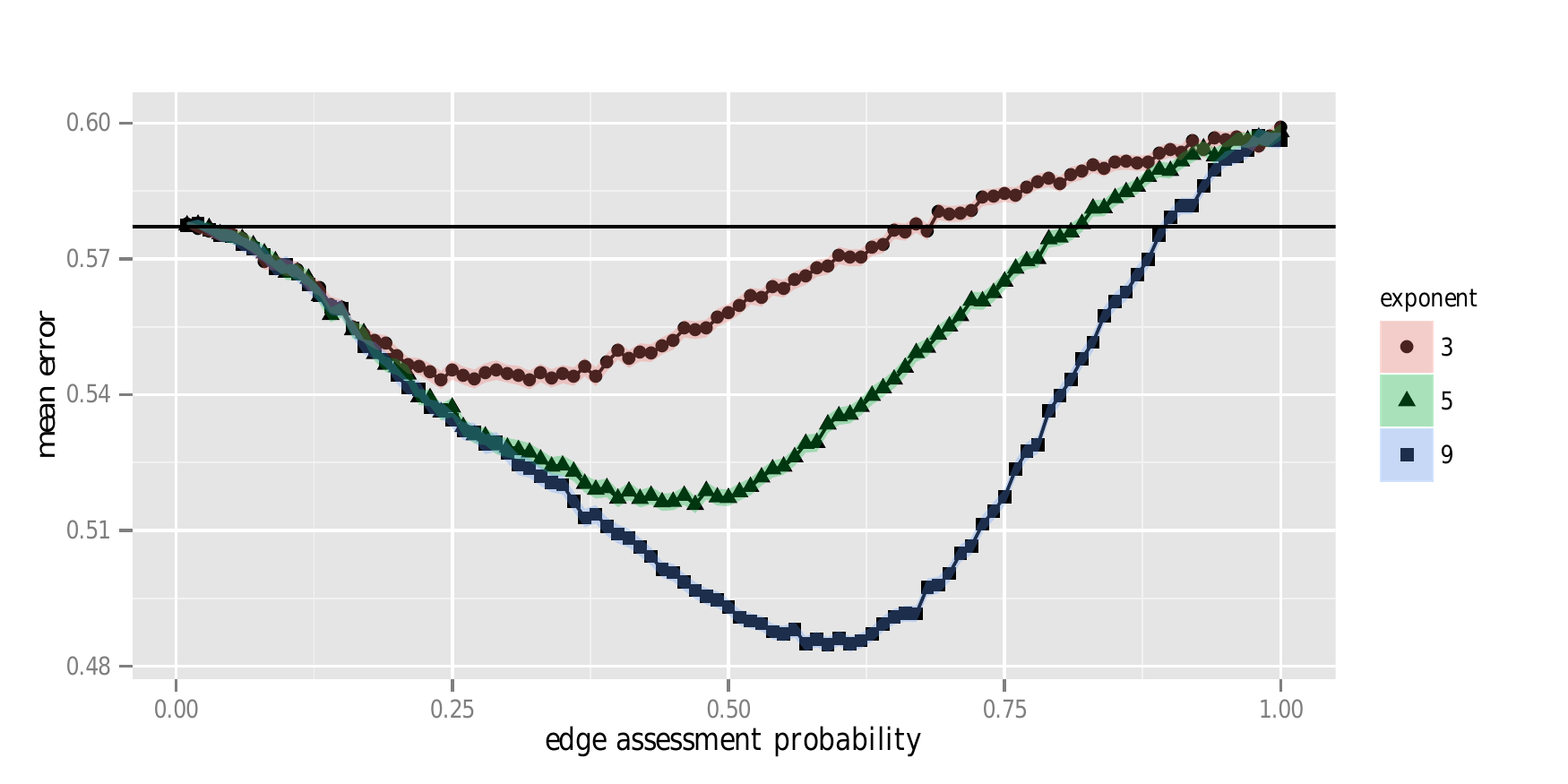}
  \end{center}
  \caption{The three curves indicate the mean leave-one-out error rate for vertex classification based on an errorful observation of the {\it C.\ Elegans} connectome as the  edge assessment probability varies with edge classification accuracy given by the equation $y=1-0.2 x^q$ for $q\in\{3,5,9\}$ (see Figure~\ref{fig:CEltri}).
  The three exponents represent different degrees of the quality/quantity tradeoff and we demonstrate that the optimal operating point and resulting performance depends on this exponent.
  The horizontal black line indicates the chance error rate for the vertex classification task.
  }
  \label{fig:CElcurve}
\end{figure}

As a further illustration of the quality/quantity trade-off for errorfully observed graphs, we will investigate this phenomenon for the {\it Caenorhabditis\ elegans} connectome \citep{varshney2011elegans,white1986structure,xu2013computer}. 
The {\it C. elegans} is a small worm and the connectome is a representation of the connections among neurons in an animal as a graph (see Appendix).
The {\it C.\ elegans} connectome consists of 302 neurons of which 279 are non-isolate somatic neurons which will be the focus of this investigation.
Two different types of connections between neurons are present in the connectome, gap junction synapses and chemical synapses, both of which form weighted connections. 
We study the gap junction connectome which forms an undirected graph and for simplicity we binarize the edges based on whether the weight of the connection is zero or non-zero. 
The 279 vertices can be divided into three classes: 118 are sensory neurons, 83 are interneurons, and 78 are motor neurons.
Hence, we will study a graph with 279 vertices and 514 edges, giving overall density $\rho=0.013$.
A depiction of the adjacency matrix for this graph is provided in Figure~\ref{fig:CEladj}.

Using the neuron class labels as the block membership function, we estimated the parameters for a stochastic blockmodel with three blocks giving
\[
  \hat{B} = \begin{bmatrix}
0.015 & 0.017 & 0.002 \\
0.017 & 0.027 & 0.012 \\
0.002 & 0.012 & 0.011 
  \end{bmatrix}
  \text{ and  }\hat{\pi} = \left[ 0.42, 0.29, 0.27  \right]'.
\]
Given estimates $\hat{B}$ and $\hat{\pi}$ we can construct the Bayes plug-in vertex classifier given by
\begin{equation}
  g(v^*) = \argmax_{k\in[3]} \hat{\pi}_k \prod_{k'=1}^3 \mathrm{Bin}(d_{k'}(v^*); n_{k'},\hat{B}_{kk'})\label{eq:plugin}
\end{equation}
where $v^*$ is the to-be-classified vertex and  $n_k=|\{v\in V\setminus\{v^*\}: Y(v)=k\}|$ for $k=1,2,3$.
As our gold standard we computed the leave-one-out error estimate in the case that the whole graph is observed giving an error rate of $0.387$.

For this data we do not have formal edge features $X(uv)$ but instead sample an errorfully observed version of the true graph by simulating the impact of choosing $\kappa$ and $\tau$.
As in our simulation example, the exact nature of the edge features and the edge classifier impact the subsequent graph inference only through their impact on the edge assessment probability and the edge classification accuracy.
The result of selection of $\kappa$ and $\tau$ given the function $h$ can be distilled as selecting two parameters: the edge assessment probability which determines the probability we will assess the edge features for a particular potential edge and is given by $h(\kappa)$, and the edge classification accuracy, the probability that the assessment will be correct which is a function of $\kappa$ and $\tau$.
Each value of $q$ represents a fixed cost curve for the graph observation procedure where as $q$ increases the overall cost increases allowing for higher quality classification for the same quantity of edge assessment.
We assume that the accuracy is identical for edges and non-edges which may at first appear\ as a simplification but as non-assessed edges are automatically assigned as non-edges this assumption is without loss of generality.

To investigate the trade-off phenomena, we simulated errorful observations of the connectome for various values of these parameters and computed the leave-one-out classification error using the Bayes plug-in classifier in Eq.~\eqref{eq:plugin}.
The results are displayed in Figure~\ref{fig:CEltri}.
Each filled area in the figure corresponds to one selection of the two parameters  and the  color indicates the mean leave-one-out classification error rate based on 1000 Monte Carlo replicates.
In the $x,y$ coordinates of figure, ie the edge-assessment-probability, edge-classification-accuracy coordinates, the point $(1,1)$ corresponds to perfect observation of the graph, the line $x=0$ corresponds to assessing zero edges and the line $y=0.5$ (not shown in the figure) corresponds to sampling an Erd\"{o}s-R\'{e}nyi graph in which non-trivial error rates are impossible.
As expected, the error rate decreases substantially as both parameters increase.

To further illustrate the quantity/quality trade-off phenomenon we considered three different curves through the space of the parameters given by the equation
$y=1-0.2 x^q$ where $x$ is the edge assessment probability $h(\kappa)$, $y$ is the edge classifiation accuracy conditional on assessment, and $q\in\{3,5,9\}$.
These curves are shown in Figure~\ref{fig:CEltri} in red, green, and blue.
Note that as $q$ increases the edge classification accuracy decreases more slowly when the edge assessment probability is small but then decreases more rapidly when the edge assessment probability is large.
In our notation, higher values of $q$ respresent less rapidly decreasing functions $h$.
This represents different degrees of the quality/quantity trade-off that may be present based on various experimental contexts.

In Figure~\ref{fig:CElcurve}, the mean leave-one-out error rates for the three curves are shown with each point again corresponding to 1000 Monte Carlo replicates.
The colored shaded areas indicate the 95\% confidence intervals for the mean leave-one-out error rates.
These curves indicate that the different values of the exponent and hence different quality/quantity trade-off regimes lead to different optimal operating points and ultimate inference task performance.
This investigation illustrates that the quantity quality trade-off phenomenon will be present even in cases that deviate substantially from the idealized setting of this paper.

\section{Conclusions} We have presented a simple model for errorfully
observed graphs derived from classifying potential edges based on
observed edge-features.  For this model, we have investigated optimal
vertex classification in the face of the quantity/quality trade-off:
more informative edge-features are more expensive, and hence the
number of potential edges that can be assessed decreases with the
quality of the edge-features.  Considering a simple vertex
classification rule, we have derived the optimal quantity/quality
operating point and demonstrated that the Bayes optimal
edge-classifier is not necessarily the optimal choice of
edge-classifier for the subsequent graph inference task.
In this section we will briefly investigate various extensions and alternative considerations to the setting presented thus far.

\subsection{Large Sample Approximation}

For sufficiently large $n$, the Binomial distributions that appear in Eq.~\eqref{cor1} can be approximated by normal distributions: $\mathrm{Bin}(n_{k}, \widetilde{B}_{kk'}) \approx \mathcal{N}(n_{k}\widetilde{B}_{kk'},n_{k}\widetilde{B}_{kk'}(1-\widetilde{B}_{kk'}) )$. 
In the simplified regime of Section~\ref{sec:demo}, we have conditioned on the fact that among the $n$ vertices with observed class labels, exactly $n/2$ are in each of the two classes and we assume $B_{11}=B_{22}>B_{12}$ resulting in the simplified form of the error rate in Eq.~\eqref{eq:simpleL}. 
We recall that the number of observed edges from vertex $v^*$ to block $k'$ is $D_k(v^*)$ which condition on $Y(v^*)=k$  is distributed as $\mathrm{Bin}(n_{k},\widetilde{B}_{kk'})$.
In this setting, the classifier in Eq.~\eqref{eq:gammaBayes} is Bayes optimal and the normal approximation to the binomial gives that, conditional on $Y(v^*)=k$, the difference $D_{2}(v^*)-D_1(v^*)$ has approximate distribution 
\begin{equation}
  \mathcal{N}\left((-1)^{k}\frac{n}{2}(\tilde{B}_{11}-\tilde{B}_{12}), \frac{n}{2}(\tilde{B}_{k2}(1-\tilde{B}_{k2})+\tilde{B}_{k1}(1-\tilde{B}_{k1}))\right).\label{eq:norm}
\end{equation}
We can use this to approximate the error rate $L_{\kappa,\tau}$ as $L^{\mathcal{N}}_{\kappa,\tau}= \Phi(-\mu/\sigma)$, where $\Phi$ is the cumulative distribution function for a standard normal and $\mu$ and $\sigma$ are the mean and standard deviation from Eq.~\eqref{eq:norm}.
As $L^{\mathcal{N}}_{\kappa,\tau}$ decreases as $\mu/\sigma$ increases, the following simplified procedure for selecting $\tau$ and $\kappa$ can be used:
\begin{align*}
(\kappa^{\mathcal{N}},\tau^{\mathcal{N}}) = \argmax_{\kappa,\tau} \quad & \mu/\sigma \\ 
\text{where}\quad & \mu = \frac{n}{2}(\tilde{B}_{11}-\tilde{B}_{12}), \\
& \sigma^2 = \frac{n}{2}(\tilde{B}_{k2}(1-\tilde{B}_{k2})+\tilde{B}_{k1}(1-\tilde{B}_{k1}))
\end{align*}
and the $2\times2$ matrix $\widetilde{B}$ is given by Eq.~\eqref{eq:1}.
A comparison of the black solid line and greed dashed line in Figure~\ref{fig:LargeSampleApprox2} illustrates the accuracy of using this approximation for selecting $\tau$ when $\kappa$ is fixed.

\subsection{Minimizing Projection Error} Spectral embedding methods
proceed by finding a low-rank latent space representation
(projection).  In the case of $SBM([n],B,\pi)$ with $rank(B)=d$,
standard results from perturbation analysis (e.g., \cite{davis70})
demonstrate that
$(\kappa^{\mathcal{P}},\tau^{\mathcal{P}}) =
\arg\min_{\kappa,\tau} \max_k (\pi \widetilde{B})_k / \lambda_d^2$,
where the numerator $(\pi \widetilde{B})_k$ is the $k^{th}$ element of
the $K$-vector $(\pi \widetilde{B})$ and the denominator $\lambda_d^2$
is the square of the $d^{th}$ largest eigenvalue of the $K \times K$
matrix $(diag(\pi) \widetilde{B})$, minimizes (with high probability)
an upper bound on the projection error.
% Experiments indicate that
% this simple indirect method yields results consistent with our exact
% solution -- that is,
% $(\widehat{\widehat{\kappa}},\widehat{\widehat{\tau}}) \approx
% (\kappa^*,\tau^*)$. 
We leave further investigations of this kind as future work and note
that for our simple demonstration case,
this approach is equivalent to our large sample approximation
$(\kappa^{\mathcal{N}},\tau^{\mathcal{N}})$.

\subsection{The Surrogate is Instructive}

The vertex classification methodology we have investigated is perhaps
the simplest nontrivial approach.  In particular, we have so far
shirked any methodology based on the common approach to general graph
inference of first embedding the graph into finite-dimensional
Euclidean space and then addressing inference therein.  The reason for
this is a clear self-indictment: we are so far mostly unable to directly
analyze the quantity/quality trade-off for statistical inference on
errorfully observed graphs in any such methodology (except
by resorting to the the asymptotic implications of a limit theorem in \cite{athreya13}).

\begin{figure}[htbp]
\begin{center}
\includegraphics[width=0.7\textwidth]{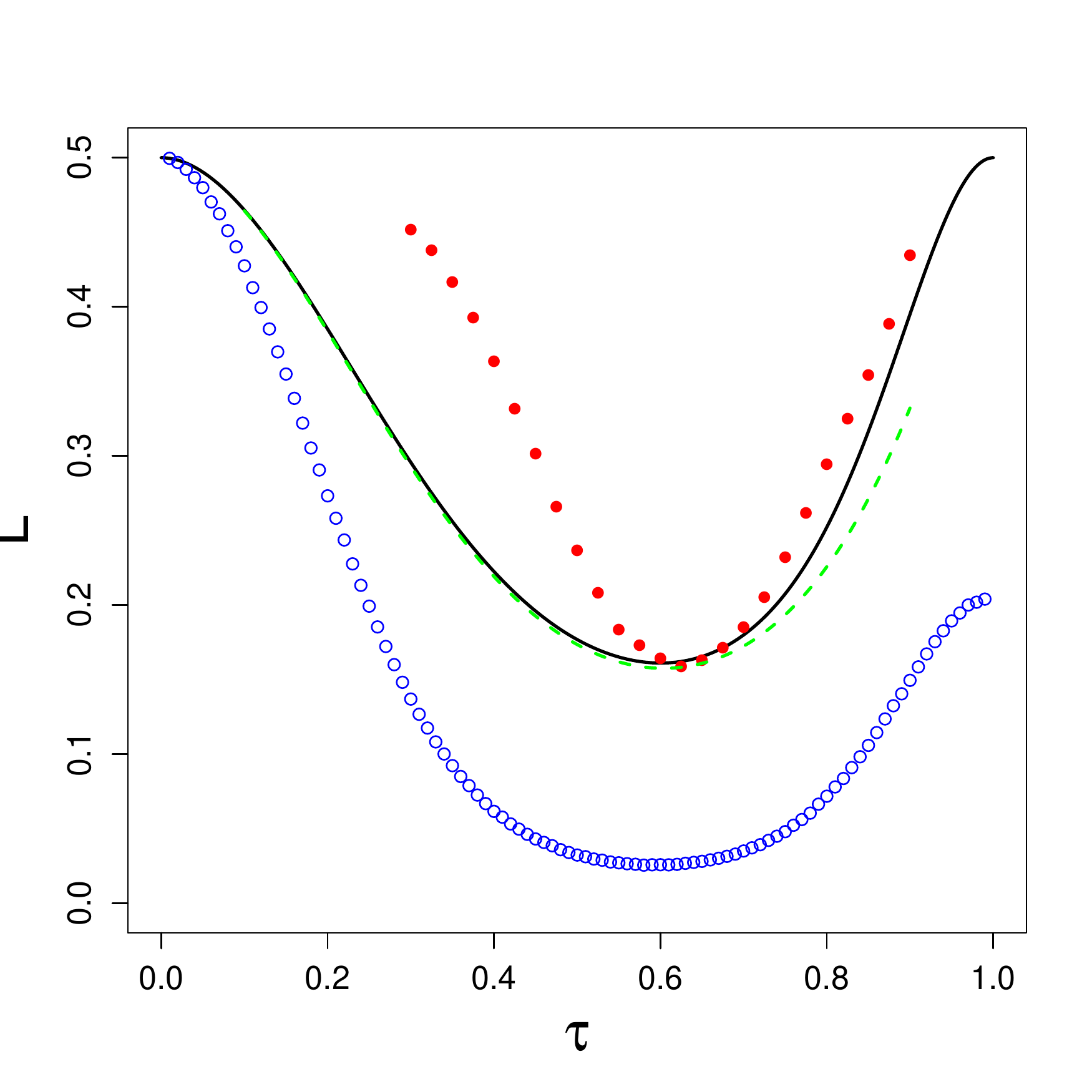}
\caption{For our demonstration setting with $\kappa^*=3.5$, we see
that the surrogate optimization is instructive regarding more
elaborate graph inference: the black solid curve is analytic
$L_{\kappa^*,\tau}$, the Bayes; the green dashed curve is the large sample normal
approximation $L^{\mathcal{N}}_{\kappa^*,\tau}$; the blue curve is
the error rate of Fisher's Linear Discriminant for points distrbuted according to the asymptotic mixture of multivariate normal distributions of the embedded points from \cite{athreya13}; the red dotted curve is
Monte Carlo $\widehat{L}^{\mathcal{P}}_{\kappa^*,\tau}$ (NB: the standard errors for the Monte Carlo are so small they are are hidden within the red dots).  Result: $\tau_{Bayes}
\approx 1/2 ; \tau^* \approx 0.600 ; \tau^{\mathcal{N}} \approx 0.604 ;
\tau^{\mathcal{P}} \approx 0.610$.  }
\label{fig:LargeSampleApprox2}
\end{center}
\end{figure}

We do, however, have a wealth of empirical evidence suggesting that
our surrogate optimization yields results that can be profitably used
to choose the $(\kappa,\tau)$ quantity/quality operating point for
these ``inference composed with embedding''
methodologies. Figure~\ref{fig:LargeSampleApprox2} presents one
illustrative empirical result supporting this
claim. Figure~\ref{fig:LargeSampleApprox2} is obtained as follows.
For our demonstration setting in \S~\ref{sec:demo}, we employ the
adjacency-spectral embedding \citep{STFP} to embed the vertices of a
graph into points in $\mathbb{R}^2$ (see also Figure~\ref{fig:YPi}). The
results in \cite{STP,tangs.:_univer} indicate that the resulting
embedding is conducive for subsequent inference, i.e., under the
latent position model of \S~\ref{sec:graph-preliminaries}, the
embeddings of the vertices converge, in the\ limit as $n \rightarrow
\infty$, to the latent positions. We then use Fisher's Linear
Discriminant \citep{DH1973} for the two-class classification in
$\Real^2$ to classify the vertices. The result, for our demonstration
setting with $\kappa^{*} = 3.5$ is given in
Figure~\ref{fig:LargeSampleApprox2}. For any fixed $(\kappa,\tau)$,
Monte Carlo yields the estimate $\widehat{L}^{\mathcal{P}}_{\kappa,\tau}$
of probability of misclassification. The number of Monte Carlo
replicates for each choice of $\kappa$ and $\tau$ is $10000$.

The blue curve in Figure~\ref{fig:LargeSampleApprox2} is a theoretical
version of the red dots based on recent results of
\cite{athreya13}. The results in \cite{STP,tangs.:_univer} are in a
sense first order results in that they demonstrated only that the
embeddings converge, in the limit, to the latent
positions. \cite{athreya13} strengthen these first order results and
by showing that for sufficiently large $n$, the embedding position of
a vertex $v$ is approximately distributed according to a multivariate normal with
mean $Z(v)$, the latent position associated with $v$, and covariance
matrix $\Sigma(v)$, with $\Sigma(v)$ converging to $0$ at the rate of
$\Theta(1/\sqrt{n})$. For a general $K \times K$ stochastic blockmodel
where each block corresponds to a class label, this means
that the embedding of a vertex $v$ with class label $k
\in \{1,2,\dots,K\}$ is distributed multivariate normal with mean
$\mu^{(k)} \in \mathbb{R}^{d}$ and covariance matrix $\Sigma^{(k)} \in
\mathbb{R}^{d \times d}$ where $d$, the rank of the matrix $B$, is the
embedding dimension.  Therefore, operating in the asymptotic regime, our
quantity/quality trade-off corresponds to finding the $\kappa, \tau$
that minimizes the error when classifying points that are distributed
as a mixture of multivariate Gaussians $\sum_{k} \pi_{k}
\phi_{\mathrm{MVN}}(\mu_{\kappa,\tau}^{(k)},
\Sigma_{\kappa,\tau}^{(k)})$ where $\mu_{\kappa,\tau}^{(k)}$ and
$\Sigma_{\kappa,\tau}^{(k)}$ can be computed explicitly, for any given
$B$ and any choice of the $\kappa$ and $\tau$ parameters. The Bayes
optimal vertex classifier, as suggested by the conjecture, corresponds
to a comparison of the densities under this multivariate normal
assumption. The blue curve in Figure~\ref{fig:LargeSampleApprox2} is
therefore the plot of the theoretical version of the red dots in
Figure~\ref{fig:LargeSampleApprox2}, where instead of performing
Fisher's Linear Discrimant for many Monte Carlo replicates, we compute
the error rate for Fisher's Linear Discriminant based on the
conjectured theoretical means and theoretical covariance matrices. The
formulae for the covariance matrices are given in \cite{athreya13}.
The resulting error rate is much lower than our other methods because 
the number of vertices is not large enough so that the asymptotic normal approximation is accurate.
On the other hand it is valuable to note that the optimal $\tau$ provided by
this method is close to the optimal $\tau$ provided by our other methods.

\subsection{The ``Missing'' Model}
\label{sec:missing-model}

The formulation we presented in Section 2 for errorfully observed
graphs $\widetilde{\GG} \sim SBM([n],\widetilde{B},\pi)$ assumes that
the potential edges not classified at all, due to the quality penalty
$h(\kappa)$, are set to 0 -- i.e., non-edge.  In fact, in many use
cases we might expect to have full knowledge of which potential edges
have been classified as non-edges and which potential edges have not
been classified at all, and it seems sensible to treat these latter as
``missing.''  (Recall that we assume that the potential edges not
classified at all are MCAR.)

This ``missing'' model is specified as in Section 2, but with two
important alterations.  First, while $\widetilde{B} =
h(\kappa)\left[(1-F_{1,\kappa}(\tau)) B + (1-F_{0,\kappa}(\tau))
(J-B)\right]$, we consider $\widetilde{B}_{MCAR} = (1-F_{1,\kappa}(\tau))
B + (1-F_{0,\kappa}(\tau)) (J-B)$.  That is, the quality penalty
$h(\kappa)$ does {\it not} impact the errorful block connectivity
probability matrix $\widetilde{B}_{MCAR}$.  Then we consider
$\widetilde{\GG}_{MCAR} \sim
s_{h(\kappa)}\left(SBM([n],\widetilde{B}_{MCAR},\pi)\right)$, where
$s_p(F_{\GG})$ for $p \in [0,1]$ and for some graph distribution
$F_{\GG}$ indicates random sampling of potential edges; the potential
edges not sampled through $s_p$ are left as missing entries in the
adjacency matrix $A$.  (Contrast this with the notionally similar
$SBM([n\sqrt{h(\kappa)}],\widetilde{B},\pi)$.)

This ``missing'' model can then be analyzed through the perspective of
imputations and inference with missing data. The literature on
statistical analysis with missing data is vast, see e.g.,
\cite{little02:_statis_analy,rubin96:_multip,reiter07}. In what
follows, we discuss briefly the analysis of this missing model and its
relationship with our quantity/quality trade-off as discussed earlier
in the paper. Our discussion is somewhat cursory, as a detailed
investigation is difficult in the scope of the paper.

We can consider the random graph $\widetilde{\GG}$ as analyzed
previously in the paper to be an instantiation of $\widetilde{\GG}_{MCAR}$ with the
missing values imputed to be 0s (non-edges). For
$\widetilde{\GG}_{MCAR}$, if we assume that the entries of the matrix
$\tilde{B}$ are known, then the Bayes optimal vertex classifier is
given as follows. For a vertex $v$, classify $v$ as being of class $1$
or $2$ depending on whether the ratio
\begin{equation*}
  \frac{ \tilde{B}_{11}^{m_1} (1 - \tilde{B}_{11})^{o_1 - m_1} \tilde{B}_{12}^{m_2}(1 -
    \tilde{B}_{12})^{o_2 - m_2} }   { \tilde{B}_{21}^{m_1} (1 - \tilde{B}_{21})^{o_1 - m_1} \tilde{B}_{22}^{m_2}(1 -
    \tilde{B}_{22})^{o_2 - m_2} }
\end{equation*} is larger or smaller than $1$, where $o_1$ and $o_2$
are the number of potential edges from $v$ to the $n_1$ and $n_2$
vertices of classes $1$ and $2$ that were classified by the edge
classifier, respectively.  We then obtain analogous optimization
results to those presented in Section 4 above. In particular, for
large $n_1 = n_2$, the likelihood in both the numerator and
denominator of the above ratio are concentrated around $o_1 \approx n
h(\kappa)/2$, $o_2 \approx nh(\kappa)/2$ ($n = n_1 + n_2$) and so the
above likelihood ratio corresponds roughly to our simple vertex
classifier in \S~\ref{sec:demo}. In particular, for
$\tilde{\mathbb{G}}_{MCAR}$ we obtain analogous optimization results
to those presented in Section 4 above. However, in general, as the
entries of the matrix $\tilde{B}$ are unknown and need to be estimated
from the data, the Bayes optimal vertex classifier for
$\widetilde{\GG}_{MCAR}$ is not as simple as our vertex classifier
$\gamma$ in \S~\ref{sec:demo}.

In summary, for our quantity/quality trade-off framework,
the difference between a complete case analysis based on labeling the missing values as
``NA'' versus analysis wherein we impute the missing values to be $0$s
(non-edges) is negligible. Of fundamental interest is therefore the quantity/quality
optimization for more elaborate imputation schemes.

\subsection{Discussion} Alas, we do not know the block connectivity
probability matrix $\widetilde{B}$ or block probability vector $\pi$.
(And of course we are not really facing a stochastic blockmodel $\ldots$ but in many applications -- for example, connectomics and social
networks -- a stochastic blockmodel can be a productive (if overly
simple) first model.)  We note that, for a given $\kappa$, one can
obtain estimates of $\widetilde{B}$ and $\pi$ from %auxillary sources.
the available $\{X(uv)\}$ and $\{Y(v)\}$, assuming a parametric form
for the class-conditional edge-feature distributions $F_{0,\kappa}$
and $F_{1,\kappa}$.  Nevertheless, our primary purpose has been to
present a foundational analysis of the quantity/quality trade-off for
errorfully observed graphs and to demonstrate the folly of fixating on
the optimization of the edge-classifier for edge-classification
performance when subsequent graph inference is the ultimate goal.

 % CEP: (for a given kappa) i can estimate Btilde from the available
 % {X(uv)} and {Y(v)}.
 %  MT: Exciting! But I presume you need to know the form for F_{0,\tau} and F_{1,\tau} ?

% Finally \dots {\it principled} ad hoc-ery! (STFP works for general RDPG: FIGURE!)

%\newpage

\section*{Appendix}

\subsection*{Connectomics}

Connectomics is a bourgeoning field in which investigators estimate
brain-graphs (connectomes) for subsequent inference tasks.  For
example, Electron Microscopy (EM) connectomics investigations explore
hypotheses of conditional independence between vertices
\citep{Bock2011}, and Magnetic Resonance (MR) connectomics often use
brain-graphs as biomarkers for phenotypic variability
\citep{Vogelstein2012}.  Regardless of the experimental modality or
subsequent inference task, connectomics investigators \emph{always}
face quantity/quality trade-offs with regard to graph inference.
These trade-offs arise in at least two stages of the analytics
pipeline: (1) data collection, (2) data analysis.  In particular, in
EM connectomics, different experimental paradigms admit different
spatial resolutions for the same imaging time \citep{Bock12}, yielding
a number of distinct $\kappa$'s.  Regardless of the chosen imaging
modality, manual, semi- or fully-automatic tracing algorithms are then
employed to infer the graph from the noisy image data
\citep{Briggman2006a}.  Each edge, therefore, can be endowed with a
confidence level, which corresponds to the edge-features of interest
described above.  Similarly, in MR connectomes, different scanner
sequences yield higher spatial resolution, but therefore reduce the
signal-to-noise ratio per voxel %\citep{Huettell2008}.
\citep{Haacke1999}.  Given the noisy MR connectomics data,
``tractography'' algorithms estimate connectivity amongst brain voxels
\citep{MRCAP11}.  Again, each edge can be endowed with a confidence.
Historically, for any connectomics investigation, the threshold for
counting an edge as ``real'' has been ad hoc, at best.  \cite{PVB}
presents a first principled treatment of this issue.  This manuscript
suggests that we can choose both $\tau$ and $\kappa$ to optimize our
subsequent inference task.

\subsection*{Social Networks}

Social network analysis is another bourgeoning field in which the data
are represented via a graph.  In this setting, vertices (actors)
represent individuals and edges (links or ties) typically represent
communication between pairs of actors.  A classic example is the Enron
email graph \citep{priebe2005scan}.  For these data, we place an edge
between a pair of actors according to whether an email was exchanged
between the pair.  Both the vertices and edges can be endowed with
complex attributes.  For example, edges may be attributed with a
word-count vector, in which each dimension corresponds to a unique
word.  The dimensionality of these attributes, however, is exceedingly
large.

We can reduce the edge attribute dimensionality via topic
modeling \citep{deerwester1990indexing, papadimitriou1998latent,
  blei2003latent}.  Topic models learn a set of ``topics'' associated
with each document (in this case, an email message).  Topic modeling
objective functions also tend to be computationally demanding.
Therefore, a number of approximations are typically employed to obtain
approximately optimal solutions that scale up to very large data,
including variants of online variational Bayes
\citep{hoffman2010online}, stochastic gradient descent
\citep{mimnostochastic}, latent factor modeling \citep{mingPFA} and
parallelization schemes \citep{scalableinference}.  Each of these
approaches makes important approximation/computation trade-offs, which
are not currently understood very well \citep{asuncion09} --
especially in terms of subsequent inference.

Recalling the Enron
email example, we may be interested in inference based on only those
email messages with certain key topics, such as sports (or insider
trading).  But assessing which emails contain the interactions of
interest is a ``Human Language Technology'' (HLT) problem.  The
computational trade-offs associated with MCMC and variational Bayesian
methods, for example, induces a quantity/quality trade-off for
assigning edge features.  Specifically, we can invest more or less HLT
time per edge, from manual investigation (humans reading the messages)
to simple keyword search: more expensive HLT will yield more accurate
topic estimation, but at the cost of fewer messages assessed, while
less expensive HLT will yield less accurate topic estimation, but for
a larger number of messages assessed.  The ability to determine the
optimal operating point for this quantity/quality trade-off for a
given computational budget will lead to superior subsequent inference
for a wide variety of social network analysis tasks.

\section*{Acknowledgments}
This work is partially funded by the National Security Science and Engineering Faculty Fellowship (NSSEFF), the Johns Hopkins University Human Language Technology Center of Excellence
(JHU HLT COE), and the 
XDATA program of the Defense Advanced Research Projects Agency
(DARPA) administered through Air Force Research Laboratory contract FA8750-12-2-0303.
We also thank the editors and
the anonymous referees for their valuables comments and critiques that greatly improved
this work.

\bibliography{errorful}
\end{document}